\pdfoutput=1

\documentclass[letterpaper, 10 pt, conference]{ieeeconf}  
\pdfoutput=1

\IEEEoverridecommandlockouts                              

\usepackage{hyperref}

\usepackage{physics}
\usepackage{amsmath,amssymb,amsfonts}

\usepackage{cite}
\usepackage{svg}
\usepackage{subfig}
\usepackage{graphicx}
\usepackage{dirtytalk}

\def\BibTeX{{\rm B\kern-.05em{\sc i\kern-.025em b}\kern-.08em
    T\kern-.1667em\lower.7ex\hbox{E}\kern-.125emX}}

\title{\LARGE \bf
Simulating User-Level Twitter Activity with XGBoost and Probabilistic Hybrid Models
}

\author{Fred Mubang $^{1}$ and Lawrence O. Hall$^{1}$
\thanks{*This work is partially supported by DARPA and Air
Force Research Laboratory via contract FA8650-18-C-7825.}
\thanks{$^{1}$Department of Computer Science ,
        University of South Florida, 4202 E Fowler Ave, Tampa, FL 33620, USA
        {\tt\small fmubang@usf.edu} {\tt\small lohall@usf.edu}}%
}

\usepackage{varioref}
\usepackage{lastpage}
\usepackage{fancyhdr}
\begin{document}

\maketitle
\thispagestyle{empty}
\pagestyle{empty}

\begin{abstract}

 The \textit{Volume-Audience-Match} simulator, or \textit{VAM} was applied to predict future activity on  Twitter related to international economic affairs.
\textit{VAM} was applied to do time-series forecasting to predict the: (1) number of total activities, (2) number of active old users, and (3) number of newly active users over the span of 24 hours from the start time of prediction. \textit{VAM} then used these volume predictions to perform user link predictions. A  user-user edge was assigned to each of the activities in the 24 future timesteps.  VAM considerably outperformed a set of baseline models in both the time series and user-assignment tasks.

\end{abstract}

\section{INTRODUCTION}
Recent research strongly suggests that social media activity can serve as an indicator for future offline events. For example, the authors of \cite{twitter-covid-bis} showed that Twitter user data could be used to predict the spatiotemporal spread of COVID-19. The authors of \cite{twitter-us2012-elec} found a strong correlation between the number of tweets mentioning each candidate in a given state, and the state's election results.

 Clearly,  more attention should be focused upon creating a simulator that can predict future social media activity at user and topic granularity. To that end, in this work we used the \textit{Volume Audience Match} Simulator, or \textit{VAM}, which was first introduced in \cite{Mubang-VAM-IEEE-TRANS}. \textit{VAM} is a machine-learning and sampling driven simulator that predicts both overall activity volume and user level activity in a given social media network.  
\textit{VAM} is comprised of 2 modules. 

The first module is the \textit{Volume Prediction Module}. This module predicts, over the next 24 hours, the future (1) activity volume time series, (2) active old user volume time series, and (3) active new user volume time series on some social media platform, for some given topic of discussion. 

The second module in \textit{VAM} is the \textit{User-Assignment Module}. This module uses the 3 time series predicted by the \textit{VP-Module}, as well as historical user-interaction information in order to predict, for a given topic, the user-to-user interactions over the next 24 hours from some start time \(T\). We tested \textit{VAM}'s predictive power on a dataset of tweets related to the China-Pakistan Economic Corridor (CPEC). 

\begin{figure*}[]
  \centering
  \includegraphics[scale=0.35]{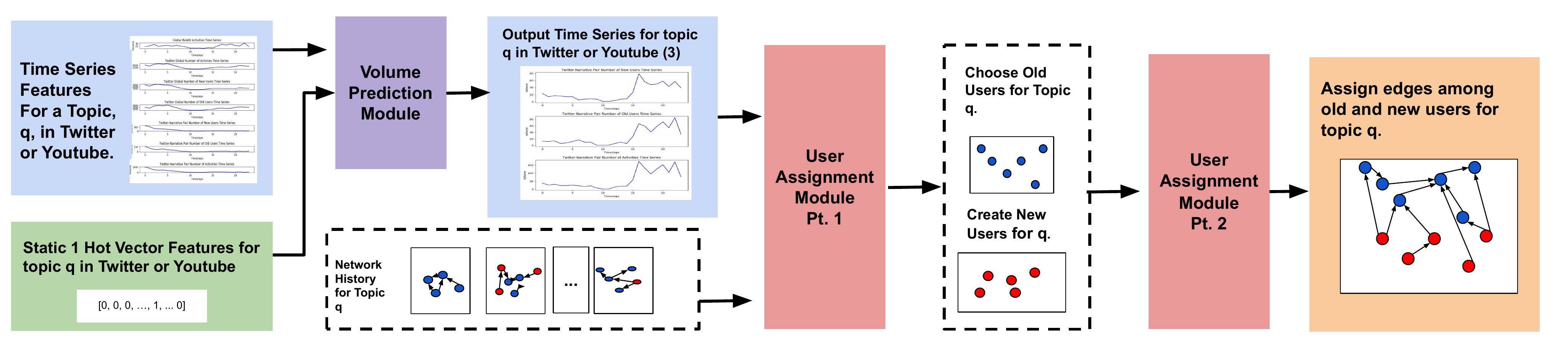}
  \caption{Framework for the Volume-Audience Match Algorithm (VAM). First, time series features and a 1 hot vector are inserted into the Volume Prediction Module. These features are used to predict the new user, old user, and activity counts. These volume counts as well as the temporal network history are then fed into the User Assignment Module to first choose the most likely old users and to create new users along with their attributes. Lastly, edges are assigned among these new and old users.}
  \label{Overview}
\end{figure*}

For the time series prediction task, we used 5 baselines and VAM outperformed all of them. These were the Persistence Model Baseline, ARIMA, ARMA, AR, and MA models\cite{box2015time}. For the user-assignment task, we used the Persistence Model Baseline (because the ARIMA-based models can only predict time series and not user-level activities). VAM outperformed this baseline in the user-assignment task as well. Figure \ref{Overview} contains a pictorial representation of \textit{VAM} \cite{Mubang-VAM-IEEE-TRANS}.

The contributions of this paper are as follows. Firstly, we show that the Volume Audience Match algorithm of \cite{Mubang-VAM-IEEE-TRANS} can be used to predict user-level activity on a dataset related to international economics (Chinese-Pakistan Economic Corridor), a dataset different than the Venezuelan Political Crisis dataset used in \cite{Mubang-VAM-IEEE-TRANS}. By using a different dataset, this lends more credence to the idea that VAM is a generalizable framework for predicting user-level activity on social media networks. Secondly, we show that VAM outperforms a multitude of baselines in the time series prediction and user-level link prediction tasks. Thirdly, similar to \cite{vam1}, we show that VAM can predict the creation of new users, unlike many previous works that only focus on the prediction of old users. Fourthly, we show that using external social media features from Reddit and YouTube can aid with predicting future Twitter activity.

\section{Motivation}

A social media topic simulator could allow governments or organizations to react to concerns of the masses more effectively and efficiently. In \cite{Mubang-VAM-IEEE-TRANS}, VAM was used to simulate future Twitter activity related to 18 Venezuelan Political Crisis topics. This domain was of interest for several reasons. For example, if there are many events or many users writing tweets about the Venezuelan \textit{ protests} topic, that could mean that there is  civil unrest taking place. Or, if many people are discussing the Venezuelan \textit{violence} topic, that could mean that many people are engaging in violent activities, or on the receiving end of such violent activities. 

In this work, we apply the VAM simulation system to another domain, which is the Chinese-Pakistan Economic Corridor (CPEC), an infrastructure initiative between China and Pakistan. There are 10 topics in this domain. If VAM could in fact, accurately predict future user-activity related to the CPEC initiative, that would allow some government or organization to have a better understanding of public opinion related to CPEC.
For example, if VAM predicts that there will be an increase in tweets related to the \textit{benefits/development/jobs} or \textit{benefits/development/roads} topics, this lets some government or corporate entity know that people may be focusing on potential benefits of the CPEC initiative such as an more jobs or better roads. 

Beyond the domain-specific applications, by applying VAM to another dataset besides the Venezuelan Political dataset of \cite{Mubang-VAM-IEEE-TRANS}, we show that perhaps VAM can serve as a general social media activity simulator, and not simply a domain-specific one.

\section{PROBLEM STATEMENTS }

As noted there are 2 problems \textit{VAM} attempts to solve, the \textit{Volume Prediction Problem} and the \textit{User-Assignment Problem}. \subsection{The Volume Prediction Problem}
The \textit{Volume Prediction Problem} is to predict the overall volume of Twitter activities. Note that we do not distinguish whether a particular action is a tweet, retweet, quote, or reply because the focus of this work is to predict the overall volume of Twitter activities. Let \(q\) be some topic of discussion on a social media platform such that \(q \in Q\), in which \(Q\) consists of all topics. Furthermore, let \(T\) be the current timestep of interest. The \textit{Volume Prediction} task is to predict 3 time series of length \(S\) between \(T + 1\) and \(T + S\). These time series, for a topic \(q\), are
the future (1) activity volume time series, which is the count of actions per time interval; (2) the active old user volume time series, which is the number of previously seen users performing an action in a time interval; and (3) the active new user volume time series, which is the number of new users that perform an action in a time interval. Note that in this work \(S=24\) in order to represent 24 hours \cite{Mubang-VAM-IEEE-TRANS}.

\subsection{The User-Assignment Problem}

Before describing the \textit{User-Assignment Problem} we must first define several terms. Let \(G\) be a sequence of temporal weighted and directed graphs such that \(G=\{G_1, G_2, ... G_T\}\). Each temporal graph, \(G_t\), can be represented as a set \((V_t,E_t) \). \(V_t\) is the set of all users that are active at time \(t\). \(E_t\) is the set of all user-to-user interactions, or links, at time \(t\). Each element of \(E_t\) is a tuple of form \((u, v, w(u, v, t)) \). \(u\) is the \textit{child } user, or the user performing an action (such as a tweet or retweet). \(v\) is the parent user, or user on the receiving end of the action. The term \(w(u, v, t))\) represents the weight of the outdegree between \(u\) and \(v\) at time \(t\) \cite{Mubang-VAM-IEEE-TRANS}.

 Now we discuss the \textit{User-Assignment Problem}. The goal is to assign a user to each activity predicted by the Volume Prediction Module, and to then assign edges between pairs of users. For tweets an edge between user A and B represents the act of user A retweeting a post by user B. 
 
 Given this information, let us say, for topic \(q\) there are 3 volume time series as discussed in the \textit{Volume Prediction Problem}. The task is now to use these volume predictions, as well as the temporal graph sequence \(G\) to predict the user-to-user interactions for topic \(q\) between \(T + 1\) and \(T + S\). This can be viewed as a temporal link prediction problem. These predicted user-user interactions are contained in a temporal graph \(\{G^{future}\}^{S}_{t=1}\) such that \(G^{future} = \{G^{future}_1, G^{future}_2, ... G^{future}_S\}\) \cite{Mubang-VAM-IEEE-TRANS}.

\section{BACKGROUND}

\subsection{The Volume-Audience-Match Simulator}

The \textit{VAM} Simulator was first discussed in the technical report \cite{Mubang-VAM-IEEE-TRANS}, which contains all details.  In this work, we focus  on the performance of \textit{VAM} on our Twitter CPEC dataset.

\subsection{General Popularity Prediction in Social Media}

Firstly, there are general \say{popularity prediction} methods which aim to predict the overall future volume of activities on a given social network, irrespective of \say{which user does what when}. This is done in \cite{madisetty-event-pop-pred} with neural networks, in \cite{fb-ts} with various statistical regression models such as Polynomial, Exponential, etc., in \cite{evently} with Hawkes Processes,  and \cite{bidoki-lstm} with LSTM neural networks.

\subsection{Decompositional User-Level Prediction in Social Media}

Next, there are methods that use a \say{decompositional user-level approach}, meaning user-level activity is predicted, however the problem is broken into several steps. The proposed models in \cite{socialcube-ieee}, \cite{renhao-github-atp}, and \cite{Horawalavithana2019MentionsSecurity} predict user activity in these two steps: (1) first, an initial model predicts the overall \say{volume} of activities, and then (2) a second model uses the predicted volume of activities as features to predict \say{which user does what when}. In \cite{socialcube-ieee}  ARIMA is used for both models and \cite{renhao-github-atp} and \cite{Horawalavithana2019MentionsSecurity} use LSTMs for both models. Note that \textit{VAM} uses a similar approach to user-level prediction, where the two main differences are that (1) VAM performs user-to-user predictions in Twitter, whereas the 3 works previously mentioned perform user-to-repository predictions in Github. Also, \cite{renhao-github-atp} and \cite{Horawalavithana2019MentionsSecurity} do not model new users at all, whereas \textit{VAM} does. New users are modeled in \cite{socialcube-ieee}  but it does not utilize a user-archetype table to model new user activity, whereas \textit{VAM} does. 

The other types of decompositional prediction methods utilize clustering. An initial model predicts the main user clusters, then a 2nd model predicts the users in each cluster \cite{saadat-github-archetypes, bidoki-coding-dyn}.

\subsection{Direct User-Level Prediction in Social Media}

The \say{direct user-level prediction} methods directly predict future user-level activity, without breaking the problem into subtasks. Embedding neural networks have been used, such as in the case of \cite{CHEN2019221} and \cite{github-renhao-cve2vec}. The works of \cite{Hernandez2020DeepLearning} and \cite{prasha-naam} use neural networks on sequences of adjacency matrices to predict user activity over time in Twitter. In \cite{garibay2020deep} the authors use a novel method that combines social science theory with a preferential attachement model. 

The authors of \cite{Blythe2019TheDS} use Bayesian, sampling, and link prediction based models to predict user-level activity. Lastly, \cite{usc2020-massive-sims} used a machine learning driven approach to predict Github user-repo pairs, as well as user cascades in Twitter and Reddit.

\subsection{General Temporal Link Prediction}
There have been several previous works on temporal link prediction algorithms. More recent embedding-based, neural-network based approaches include \textit{dyngraph2vec} \cite{dg2vec}, and \textit{tNodeEmbed} from \cite{tnode}. However, the issue with embedding methods is that they are computationally expensive in terms of training time and space.

There are matrix factorization methods used in  \cite{temp-lp-dunlavy}, \cite{non-neg-mf}, and \cite{gao-temp-lp}. And finally, probabilistic methods, such as \cite{non-para-lp-sarkar} and \cite{ahmed-ts-rand-walk}. However, the problem with these approaches is that they are computationally expensive in terms of time and space. Lastly, none of these approaches can predict the growth of new users, which is important for certain social networks in which activity is strongly driven by new users.

\section{DATA COLLECTION}

Data was collected and anonymized by  Leidos. Annotators and subject matter experts (SMEs) worked together to annotate an initial set of 4,997 tweet and YouTube comments. These posts were related to 21 different topics, which are in the supplemental materials \cite{vam-suppl}. There is a table that shows the Weighted Average Inner-Annotator agreements on each of these topics. All topics are related to the Chinese-Pakistan Economic Corridor. The time period was from April 2, 2020 to August 31, 2020.  

A BERT model \cite{bert} was  trained and tested on this annotated data with a train/test split of 0.85 to 0.15. The F1 scores per topic are shown in the supplemental materials \cite{vam-suppl}. There was a wide range of F1 scores, with the highest being 0.97 and the lowest being 0. As a result, in order to avoid having an overtly \say{noisy} dataset, we only chose topics for our final Twitter dataset that had a Weighted Average Inner-Annotator Agreement of 0.8 or higher, and a BERT F1 score of 0.7 or higher. By doing this, we ended up with 10 topics.

This BERT model was then used to label topics for 3,166,842 Twitter posts (tweets/retweets/quotes/replies) and  5,620 YouTube posts (videos and comments). Table \ref{tab:YouTube-twitter-topic-counts} shows the counts of the Twitter and YouTube posts per topic. BERT was not applied to the Reddit data, so the Reddit data used as additional features in this work is not split by topics. 

The supplemental materials \cite{vam-suppl} contain the node and edge counts of each of the 10 Twitter networks. The largest network in terms of nodes is the \textit{controversies/china/border} network with 443,666 nodes. The smallest network in terms of nodes is the \textit{controversies/pakistan/students} network, with 10,650 nodes.

Lastly, the supplemental materials \cite{vam-suppl} contain a table showing the average hourly proportion of new to old users in the Twitter dataset. As shown in the table, for some topics, there is a particularly high frequency of average new users per hour. For example, in controversies/china/uighur, on average, every hour 78.72\% of the active users were new and 21.28\% were old. Topics such as this are the reason we aim to use \textit{VAM} to predict both new and old user activity, unlike most previous works that only focus on old/previous user activity prediction.

\begin{table}[]
\begin{tabular}{|ccc|}
\hline
\multicolumn{3}{|c|}{\textbf{Twitter and YouTube Topic Counts}}                                \\ \hline
\multicolumn{1}{|c|}{\textbf{Topic}} & \multicolumn{1}{c|}{\textbf{Twitter Counts}} & \textbf{Youtube Counts} \\ \hline
\multicolumn{1}{|c|}{controversies/china/border}      & \multicolumn{1}{c|}{1,509,000} & 1,081 \\ \hline
\multicolumn{1}{|c|}{controversies/pakistan/baloch}   & \multicolumn{1}{c|}{344,289}   & 856   \\ \hline
\multicolumn{1}{|c|}{opposition/propaganda}           & \multicolumn{1}{c|}{309,378}   & 455   \\ \hline
\multicolumn{1}{|c|}{benefits/development/roads}      & \multicolumn{1}{c|}{189,082}   & 937   \\ \hline
\multicolumn{1}{|c|}{leadership/sharif}               & \multicolumn{1}{c|}{185,851}   & 648   \\ \hline
\multicolumn{1}{|c|}{controversies/china/uighur}      & \multicolumn{1}{c|}{173,431}   & 440   \\ \hline
\multicolumn{1}{|c|}{benefits/development/energy}     & \multicolumn{1}{c|}{160,874}   & 436   \\ \hline
\multicolumn{1}{|c|}{leadership/bajwa}                & \multicolumn{1}{c|}{144,277}   & 494   \\ \hline
\multicolumn{1}{|c|}{benefits/jobs}                   & \multicolumn{1}{c|}{112,769}   & 267   \\ \hline
\multicolumn{1}{|c|}{controversies/pakistan/students} & \multicolumn{1}{c|}{37,891}    & 6     \\ \hline
\end{tabular}
\caption{Twitter and YouTube post counts per topic. Twitter counts refer to tweets, retweets, quotes, and replies. YouTube posts refer to videos and comments.}
\label{tab:YouTube-twitter-topic-counts}
\end{table}

\section{VOLUME PREDICTION METHODOLOGY}

\subsection{Data Processing}
Our training period was from April 2, 2020 to August 10, 2020 (4 months). The validation period was August 11 to August 17th, 2020 (1 week). Lastly, the test period was August 18, 2020 to August 31, 2020 (2 weeks).

Each sample represents a \textit{topic-timestep} pair. The input features represent multiple time series leading up to a given timestep of interest \(T\). The different possible time series used for features are shown in Table \ref{tab:time-series-table}. Also, a 1 hot vector of size 10 was used to indicate which topic each sample represented. 

Table \ref{tab:model-ft-sizes} shows the feature sizes for each model trained. The \textit{model} column shows the name of the model. The abbreviation represents the platform features used to train the particular model. \say{T}, \say{Y}, and \say{R} represent Twitter, YouTube, and Reddit respectively. The numbers represent the hourly length of the time series input to each model. However, note that the 3 output time series of each model are each of length 24 in order to maintain consistency in evaluation. For example, the \textit{VAM-TR-72} model is a model trained on Twitter and Reddit time series that are all of length 72. Using Table \ref{tab:time-series-table} these time series indices would be 1-3, 7-9, and 13, or 7 different time series. Also recall, the 10 static features (for the 1 hot vector). So in total, this model had 7*72 + 10 = 514 features, as shown in the table. 

There were 31,210 training samples used for each model, 1,450 validation samples, and 140 test samples. There are 140 test samples because of 10 topics and 14 days for testing. However, for training and validation, we wanted to generate as many samples as possible so our models had adequate data. So, for those datasets, we created samples by creating \say{days} both in terms of hour and day. We call this a \say{sliding window data generation} approach, similar to \cite{Mubang-VAM-IEEE-TRANS}.

We trained 12 different VAM models. Each model was trained on a different combination of platform features which were some combination of Twitter, Reddit, and YouTube. The time series features used for each platform are shown in Table \ref{tab:time-series-table}. The names of the different models used are shown in Table \ref{tab:vam-vol-ranks}. Furthermore, we also used different \textit{volume lookback factors} (\(L^{vol}\)). The \(L^{vol}\) parameter determines the length of each time series described in Table \ref{tab:time-series-table}. For example, the \textit{VAM-TRY-24} model was the model trained on Twitter, Reddit, and YouTube time series, all of length 24.

\subsection{XGBoost}
\textit{VAM}'s \textit{Volume Prediction module}, which we call \(\Phi\), is comprised of multiple XGBoost models. It takes an input vector, \(\vb{x}\) and produces a matrix, \(\boldsymbol{\hat{Y}} \in \mathbb{R}^{3 \times S}\). In other words, \(\Phi(\vb{x})=\boldsymbol{\hat{Y}}\).  Each row of this matrix represents one of the 3 volume time series (actions, new users, old users). Each column represents a timestep between T + 1 and \(T + S\), with \(S=24\) hours in our experiments. As a result, there are 72 XGBoost models contained within the Volume Prediction Module \(\Phi\), each one \say{specializing} on an hour-output-type pair (e.g. number of new users in hour 1, or number of activities at hour 18, etc. ). For more details see \cite{Mubang-VAM-IEEE-TRANS}.

\subsection{Parameter Selection}

Similar to \cite{Mubang-VAM-IEEE-TRANS }, we used the \textit{XGBoost} \cite{xgboost} and \textit{sk-learn} \cite{scikit-learn} libraries to create our XGBoost models.  The subsample frequency, gamma, and L1 regularization parameters were set to 1, 0, and 0 respectively.  A grid search over a pool of candidate values was done for other parameters using the   validation set. For the \textit{column sample frequency}, the candidate values were 0.6, 0.8, and 1. For the \textit{number of trees} parameter, the candidate values were 100 and 200. For the \textit{learning rate}, the values were 0.1 and 0.2. For \textit{L2 Regularization}, the values were 0.2 and 1. Lastly, for \textit{maximum tree depth}, the values were 5 and 7.

Mean Squared Error was the loss function and log normalization was used.

\begin{table}[h!]
\centering
\begin{tabular}{|l|l|}
\hline
\textbf{\begin{tabular}[c]{@{}l@{}}Time \\Series \\ Index \end{tabular}} & \textbf{Time Series Description}                                                                                \\ \hline
\textbf{1}                                                                    & \begin{tabular}[c]{@{}l@{}}New user volume time series for a given topic in Twitter.\end{tabular}  \\ \hline
\textbf{2}                                                                    & \begin{tabular}[c]{@{}l@{}}Old user volume time series for a given topic in Twitter.\end{tabular}  \\ \hline
\textbf{3}                                                                    & \begin{tabular}[c]{@{}l@{}}Activity volume time series for a given topic in Twitter.\end{tabular} \\ \hline
\textbf{4}                                                                    & \begin{tabular}[c]{@{}l@{}}New user volume time series for a given topic in YouTube.\end{tabular}  \\ \hline
\textbf{5}                                                                    & \begin{tabular}[c]{@{}l@{}}Old user time series for a given topic in YouTube.\end{tabular}  \\ \hline
\textbf{6}                                                                    & \begin{tabular}[c]{@{}l@{}}Activity volume time series for a given topic in YouTube.\end{tabular} \\ \hline
\textbf{7}                                                                    & \begin{tabular}[c]{@{}l@{}}Activity volume time series across all topics in Twitter.\end{tabular}     \\ \hline
\textbf{8}                                                                    & \begin{tabular}[c]{@{}l@{}}New user volume time series across all topics in Twitter.\end{tabular}      \\ \hline
\textbf{9}                                                                    & \begin{tabular}[c]{@{}l@{}}Old user volume time series across all topics in Twitter.\end{tabular}      \\ \hline
\textbf{10}                                                                   & \begin{tabular}[c]{@{}l@{}}Activity volume across  all topics in YouTube.\end{tabular}     \\ \hline
\textbf{11}                                                                   & \begin{tabular}[c]{@{}l@{}}New user volume time series across all topics in YouTube.\end{tabular}      \\ \hline
\textbf{12}                                                                   & \begin{tabular}[c]{@{}l@{}}Old user volume time series across all topics in YouTube.\end{tabular}      \\ \hline 
\textbf{13}                                                                   & \begin{tabular}[c]{@{}l@{}}Activity volume time series in Reddit.\end{tabular}                     \\ \hline
\end{tabular}
\caption{All possible time series feature categories.}
\label{tab:time-series-table}
\end{table}

\begin{table}[h!]
\centering
\begin{tabular}{|cc|}
\hline
\multicolumn{2}{|c|}{\textbf{Model Input Feature Sizes}} \\ \hline
\multicolumn{1}{|c|}{\textbf{Model}} & \textbf{Features} \\ \hline
\multicolumn{1}{|c|}{VAM-TR-72}      & 514               \\ \hline
\multicolumn{1}{|c|}{VAM-TY-72}      & 874               \\ \hline
\multicolumn{1}{|c|}{VAM-TRY-48}     & 634               \\ \hline
\multicolumn{1}{|c|}{VAM-TR-48}      & 346               \\ \hline
\multicolumn{1}{|c|}{VAM-TRY-72}     & 946               \\ \hline
\multicolumn{1}{|c|}{VAM-T-72}       & 442               \\ \hline
\multicolumn{1}{|c|}{VAM-TY-48}      & 586               \\ \hline
\multicolumn{1}{|c|}{VAM-T-48}       & 298               \\ \hline
\multicolumn{1}{|c|}{VAM-TR-24}      & 178               \\ \hline
\multicolumn{1}{|c|}{VAM-TRY-24}     & 322               \\ \hline
\multicolumn{1}{|c|}{VAM-TY-24}      & 298               \\ \hline
\multicolumn{1}{|c|}{VAM-T-24}       & 154               \\ \hline
\end{tabular}
\caption{Twitter volume model input sizes.}
\label{tab:model-ft-sizes}
\end{table}

\section{VOLUME PREDICTION RESULTS}

\subsection{Metrics Used}

In order to properly assess VAM's predictive power in the time series prediction task, various metrics were used. We used RMSE and MAE metrics in order to assess how well VAM could predict time series in terms of volume and exact timing. 

Predicting the exact timing of a time series is a difficult task. It is possible for a model to approximate the overall \say{shape} of a time series, while not correctly predicting the number of events or exact temporal pattern. In order to account for this phenomenon, we also use the \text{Normalized RMSE} metric. It is calculated in the following way. The ground truth time series and simulated time series are both converted into cumulative time series. Each time series is then divided by its respective maximum value. The result is 2 time series whose values range from 0 to 1. Finally, the standard RMSE metric is applied to these normalized time series.

In order to measure VAM's accuracy in terms of pure volume of events, without regard to temporal pattern, we used the Symmetric Absolute Percentage Error, or \textit{S-APE}. This measures how accurate the total number of events was for each model, without regard to the temporal pattern. The formula is as follows. Let \(F\) be the forecast time series, and let \(A\) be the actual time series:

\[ SAPE = \frac{\abs{sum(F) - sum(A)}}{sum(F) + sum(A) } * 100\%  \]

The last 2 metrics were used in order to measure how well the volatility of a predicted time series matches that of the ground truth. These metrics are \textit{Volatility Error} (VE) and \textit{Skewness Error} (SkE). The Volatility Error is measured by calculating the absolute difference between the actual and predicted time series' standard deviations. The Skewness Error is measured by calculating the absolute difference between the actual and predicted time series' skewness. The skewness statistic used in this work utilizes the adjusted Fisher-Pearson standardized moment coefficient  \cite{doane-meas-skewness}.

\subsection{Baselines Used}

We compared VAM to 5 baseline models, which are the Persistence Baseline, ARIMA, ARMA, AR, and MA models \cite{box2015time}. Firstly, we used the Persistence Baseline which is defined as follows. Let \(T\) be the current timestep of interest, and let \(S\) is the length of the desired predicted time series. The Persistence Model predicts the time series for period \(T +1\) to \(T + S\) by moving forward, or \say{shifting forward} the time series that spans period \(T -S \) to \(T\). The underlying assumption of this model is that the immediate future of the time series will simply replicate its immediate past. Within the realm of social media time series prediction, this baseline has been used in \cite{prasha-naam, Hernandez2020DeepLearning, socialcube}. 

The Auto-Regressive Integrated Moving Average model (ARIMA) and its variants (ARMA, AR, and MA) are widely used statistical models and, hence, used for comparison. Furthermore, these models have even been used as the basis for some recent prediction approaches such as in the case of \cite{socialcube-ieee} and \cite{bidoki-coding-dyn}. The ARMA, AR, and MA models are variants of ARIMA depending on what the \textit{p, d,} and \textit{q} parameters are set to. The term \textit{p} represents the number of autoregressive terms. \textit{d} is the number of differences to be performed for stationarity in the time series. Lastly, \textit{q} is the number of lagged forecast errors to be used in the prediction equation \cite{box2015time}. For more details regarding how the ARIMA model and its variants had parameters set, refer to the supplemental materials \cite{vam-suppl}.

\subsection{Metric Results}

Table \ref{tab:vam-vol-ranks} contains the 6 metric results for the 12 VAM models and 5 baselines. Since it is difficult to evaluate which model is best from looking at 6 different raw metric scores, we created one general metric called the \textit{Overall Normalized Metric Error (ONME)}. It is calculated in the following way. For each metric column, the 17 values (for 17 models) are normalized between 0 and 1 by dividing each value by the sum of the 17 metric scores. By doing this, one then creates 6 different scores for each model, all between 0 and 1, although no score is exactly 0 or 1, as one can see. We used this normalization method to show that the best model still had some error. Finally, these 6 scores for each model are averaged, creating the ONME score. The models in this table are ranked from best to worst in terms of ONME. Lower ONME is better.

Note that ARMA was the best baseline model because it had the lowest Overall Normalized Metric Error out of all 5 baselines. We wanted to know how well each model performed in comparison to this baseline. So, we also created the \textit{ONME Percent Improvement From Best Baseline Metric (PIFBB)}, in a similar fashion to \cite{Mubang-VAM-IEEE-TRANS}. It is calculated in the following way: \[PIMFBB =  100\% * \frac{Best Baseline Error - Model Error}{Best Baseline Error}  \]. 

The upper bound of \(PIMFBB\) is 100\%, which occurs if a model's \textit{PIMFBB} is 0. This is clearly the best possible result. The lower bound for \(\textit{ONME}\) is negative infinity because any given model could potentially perform infinitely worse than the best baseline.

According to Table \ref{tab:vam-vol-ranks}, the best model was the \textit{VAM-TR-72} model. This was the VAM model trained on both Twitter and Reddit features with a lookback factor of 72. Its ONME Percent Improvement From the Best Baseline (ARMA) was 16.92\%. 
It is noteworthy that the best 5 models all used Reddit and YouTube features in addition to Twitter features in order to predict the Twitter time series. This suggests that external platform features from Reddit and YouTube can be helpful in predicting future events on Twitter. Furthermore, the 4 worst VAM models all had lookback factors of 24, suggesting that longer lookback periods of 48 or 72 are more helpful for accurate Twitter time series prediction.

Lastly, while ARMA was the best baseline model, it is noteworthy that the seemingly simplistic Persistence Baseline was the 2nd best baseline, ahead of the well-known ARIMA model. The Persistence Baseline's PIFBB was -0.04\%, while ARIMA's was -4.62\% (higher is better). This is noteworthy because the Persistence Baseline is simply created by \say{shifting ahead} the events from the past to the future time frame. 

\subsection{Per Topic Analysis}

\begin{figure*}[t!]
	\centering
\begin{tabular}{ccc}

		\subfloat[RMSE (Lower is Better)]{
         \includegraphics[scale=0.4]{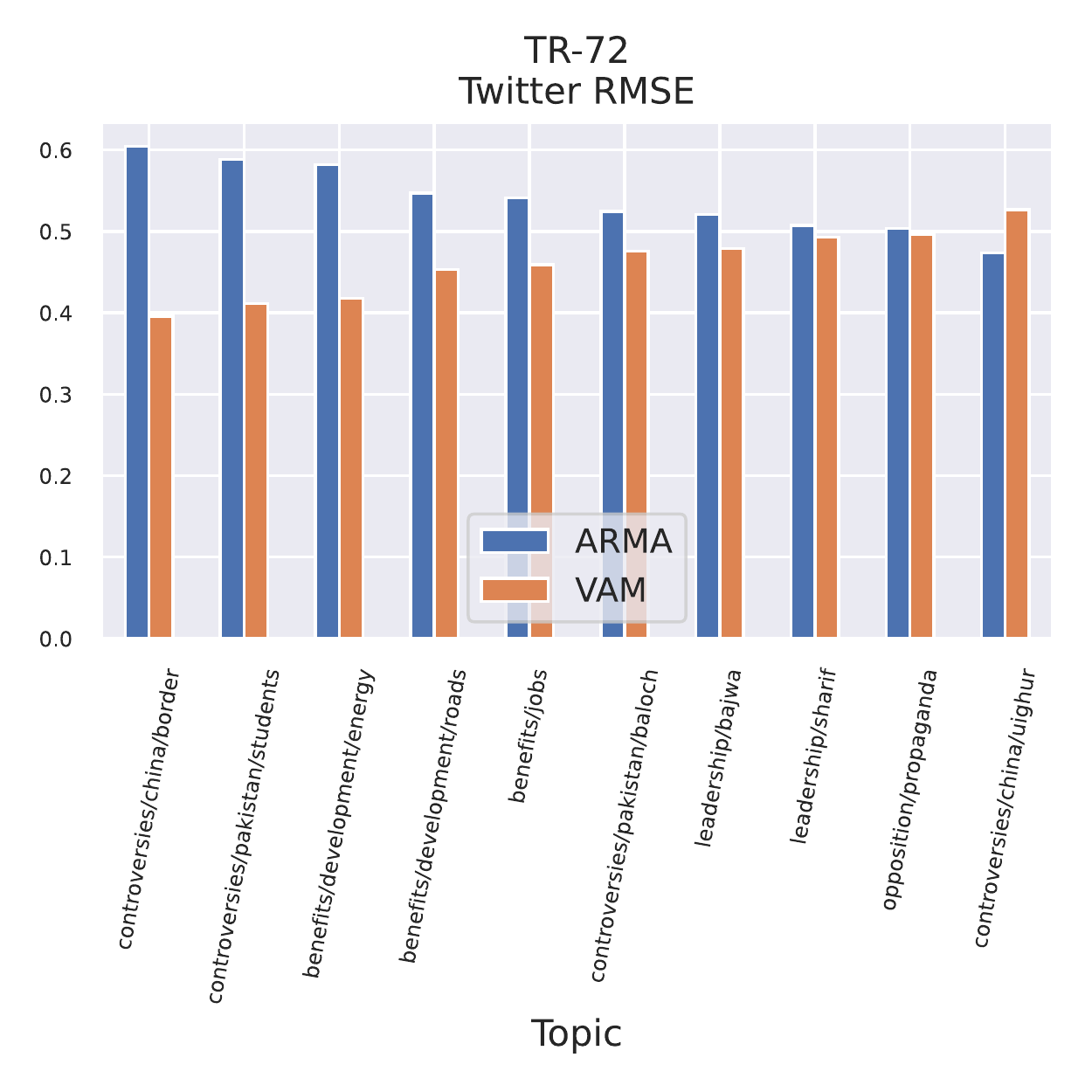} 
		    \label{fig:twitter-VAM-TR-96-RMSE-results}
	}
			\subfloat[MAE (Lower is Better)]{
         \includegraphics[scale=0.4]{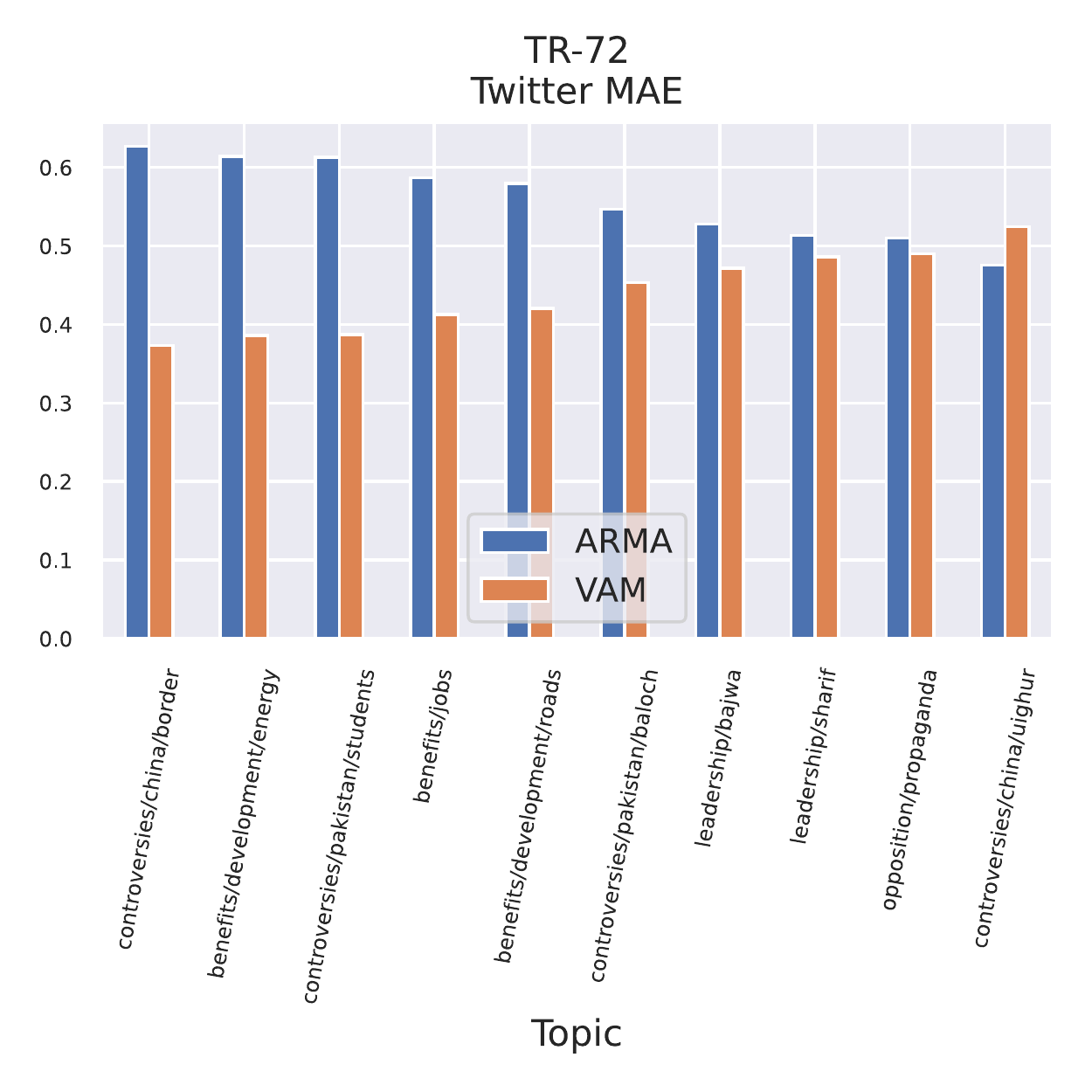} 
		    \label{fig:twitter-VAM-TR-96-MAE-results}
	}
			\subfloat[NC-RMSE (Lower is Better)]{
         \includegraphics[scale=0.4]{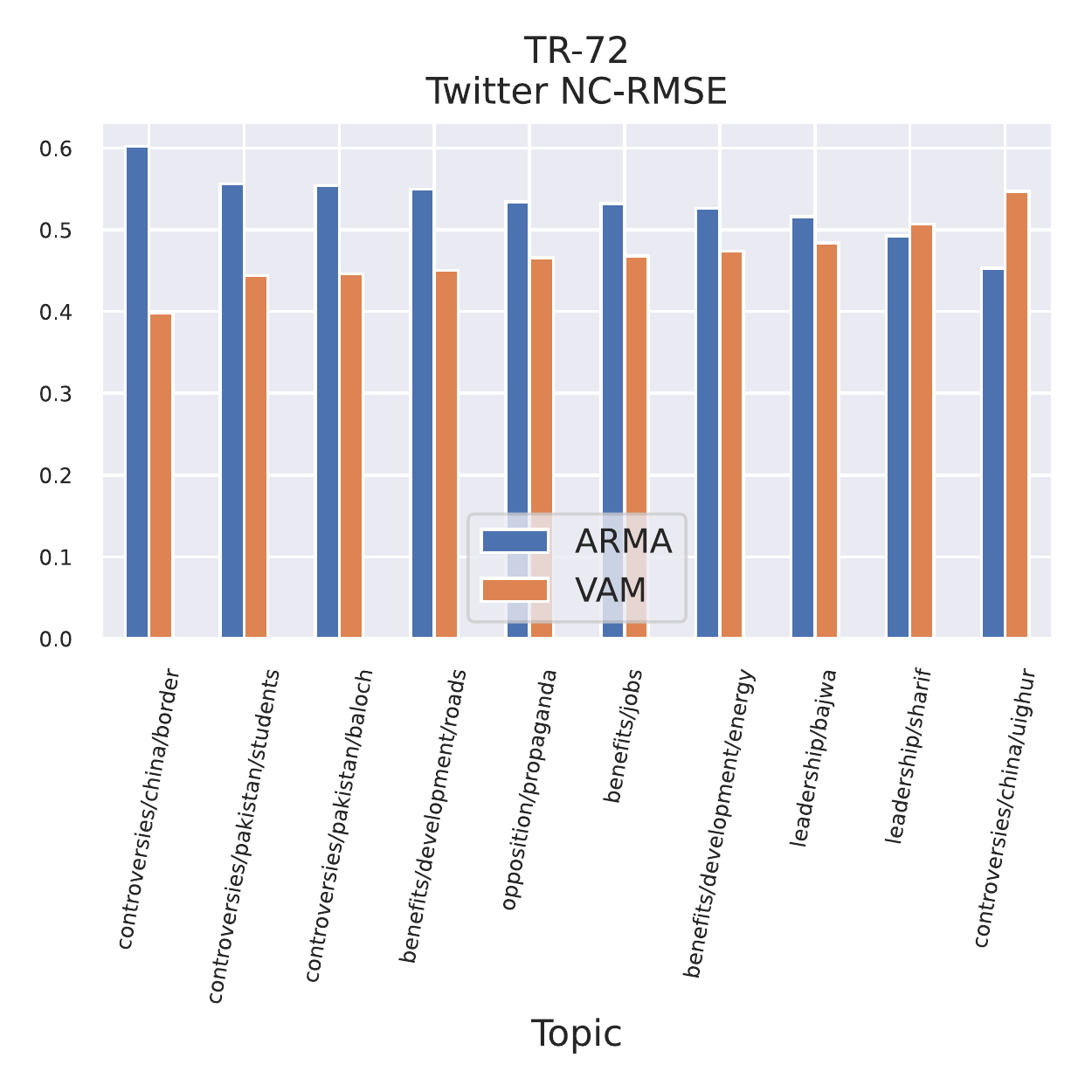} 
		    \label{fig:twitter-VAM-TR-96-NC-RMSE-results}
	}
	
	\\
	
			\subfloat[S-APE (Lower is Better)]{
         \includegraphics[scale=0.4]{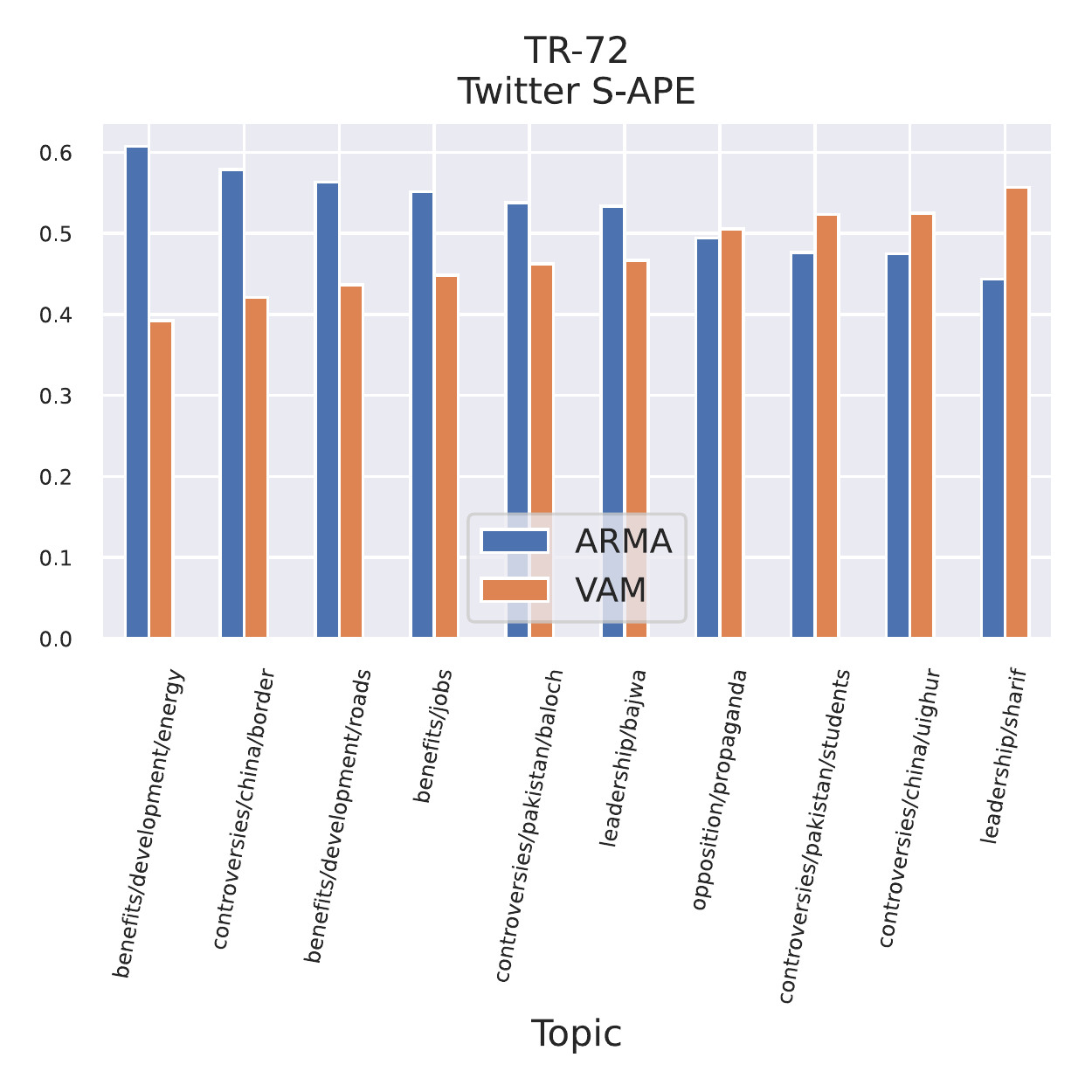} 
		    \label{fig:twitter-VAM-TR-96-S-APE-results}
	}
	
			\subfloat[Skewness Error (Lower is Better)]{
         \includegraphics[scale=0.4]{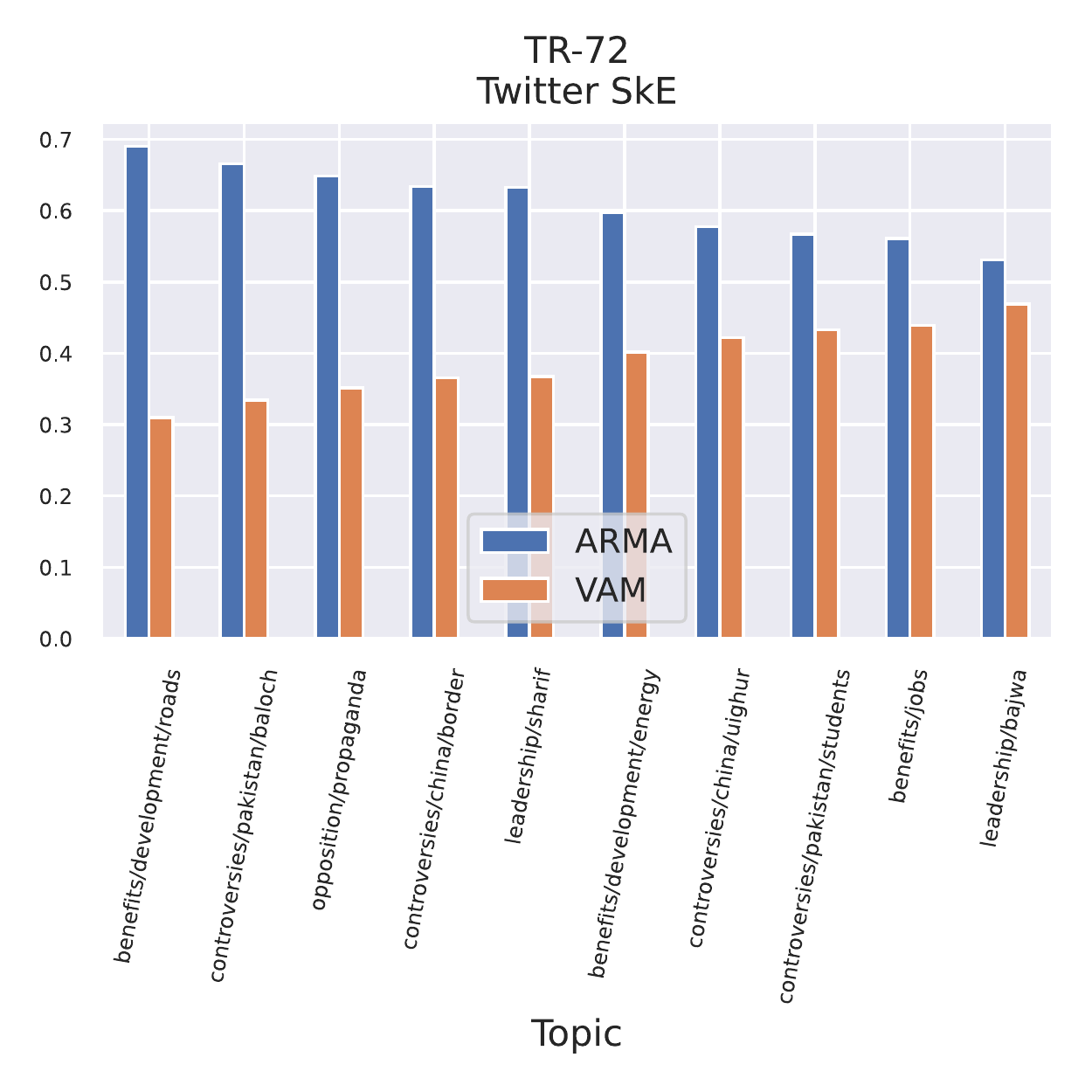} 
		    \label{fig:twitter-VAM-TR-96-SkE-results}
	}

		\subfloat[Volatility Error (Lower is Better)]{
         \includegraphics[scale=0.4]{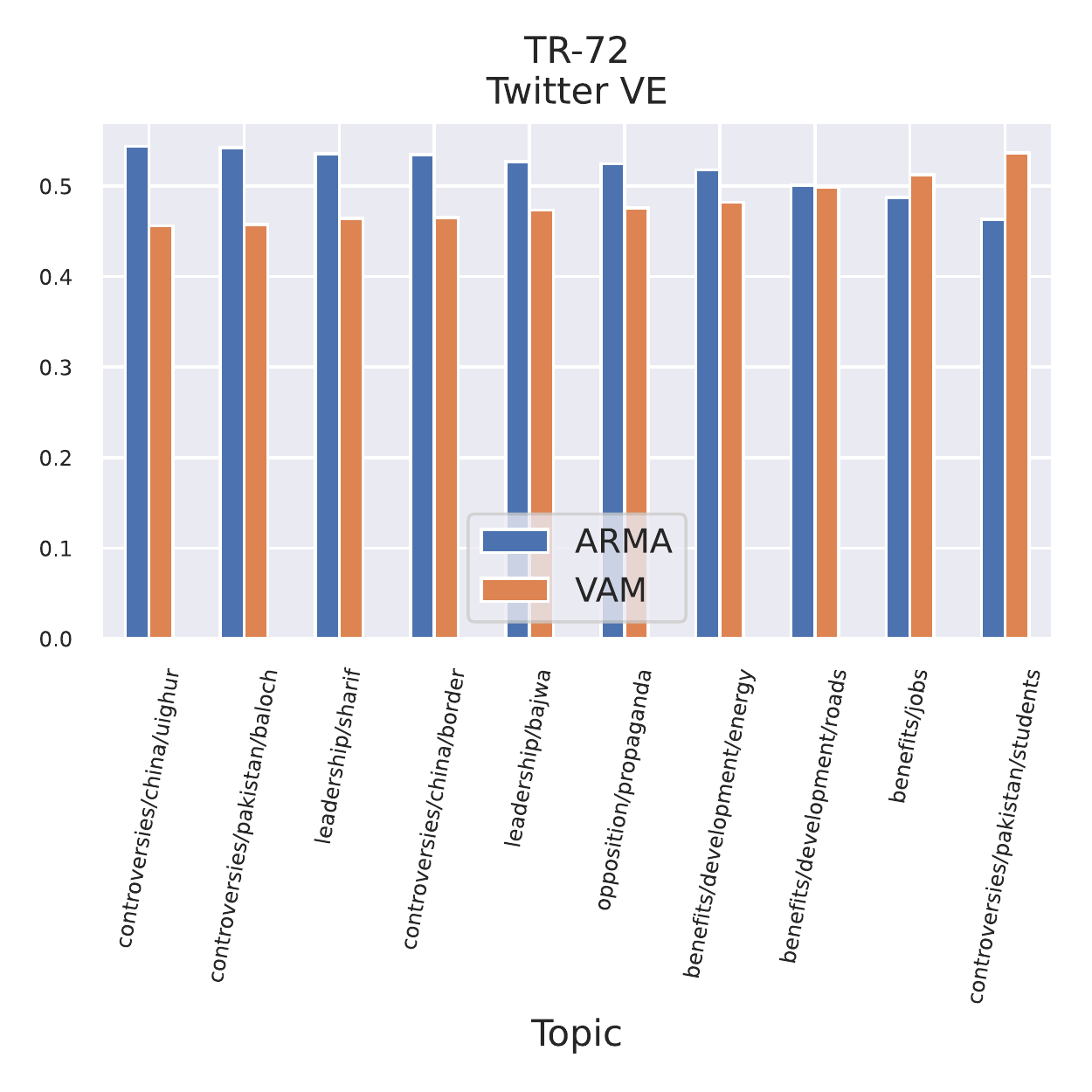} 
		    \label{fig:twitter-VAM-TR-96-VE-results}
	}

\end{tabular}
\caption{VAM (orange) vs. the best baseline from Table \ref{tab:vam-vol-ranks}, ARMA, (blue) across various topics. The metric results per topic for both models were normalized between 0 and 1 for easier visualization.  }
\label{fig:twitter-vol-barplots}
\end{figure*}

Figure \ref{fig:twitter-vol-barplots} contains bar plots showing how the VAM-TR-72 model performed against ARMA on a per-topic basis for each metric. The orange bars represent the VAM ONME scores and the blue bars represent the ARMA ONME scores (lower is better). 
For RMSE, the VAM model outperformed the ARMA model on 8 out of 10 topics. For MAE, VAM won 9 out of 10 times. For NC-RMSE, VAM won 8 out of 10 times. For S-APE, VAM won 6 out of 10 times. For Skewness Error, VAM won 10 out of 10 times. Lastly, for Volatility Error, VAM won 8 out of 10 times. 

\begin{figure*}[t!]
	\centering
\begin{tabular}{cc}
		\subfloat[Benefits/Dev/Energy]{
         \includegraphics[scale=0.16]{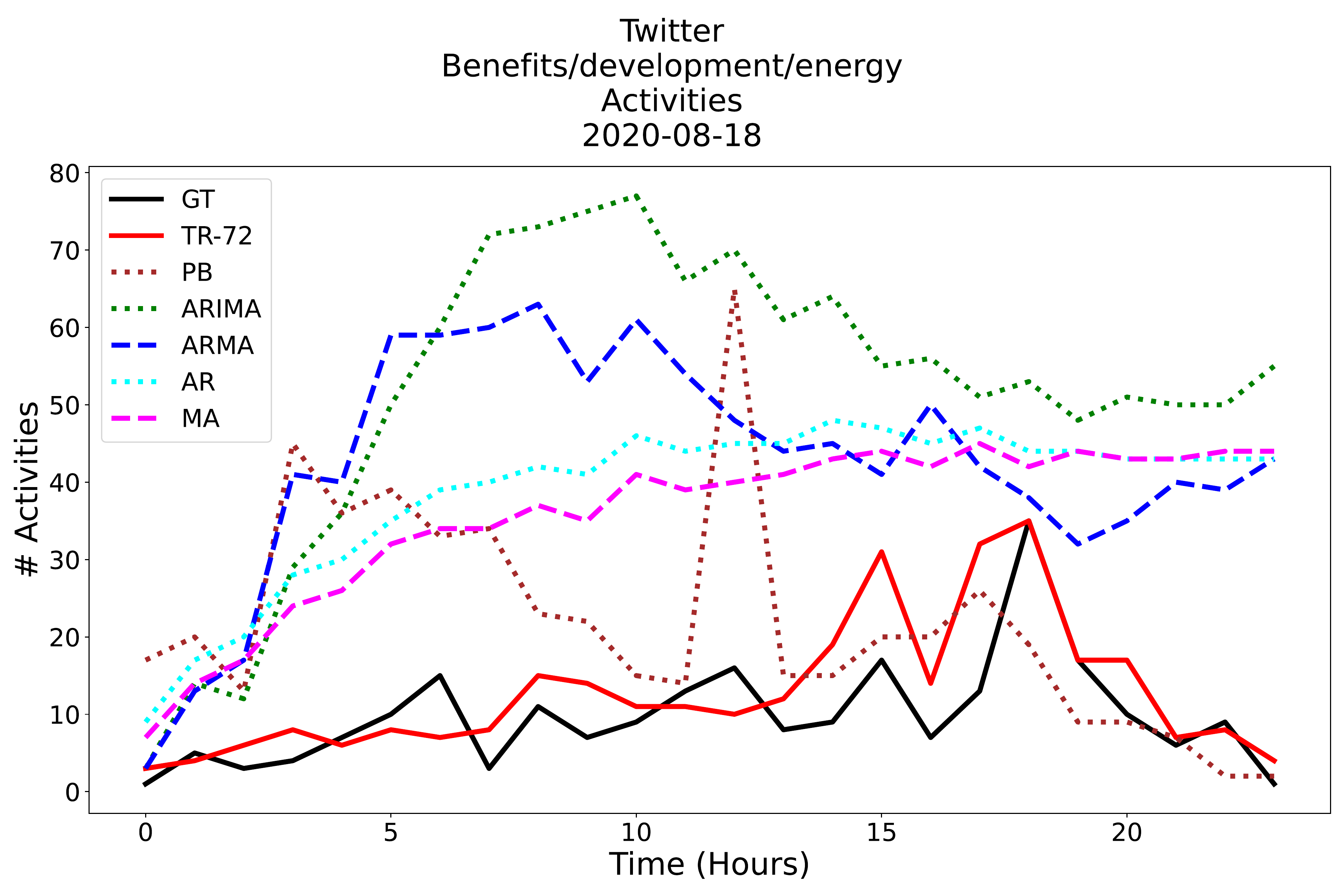} 
		    \label{fig:ts-plot1}
	}

		\subfloat[Benefits/Jobs]{
         \includegraphics[scale=0.16]{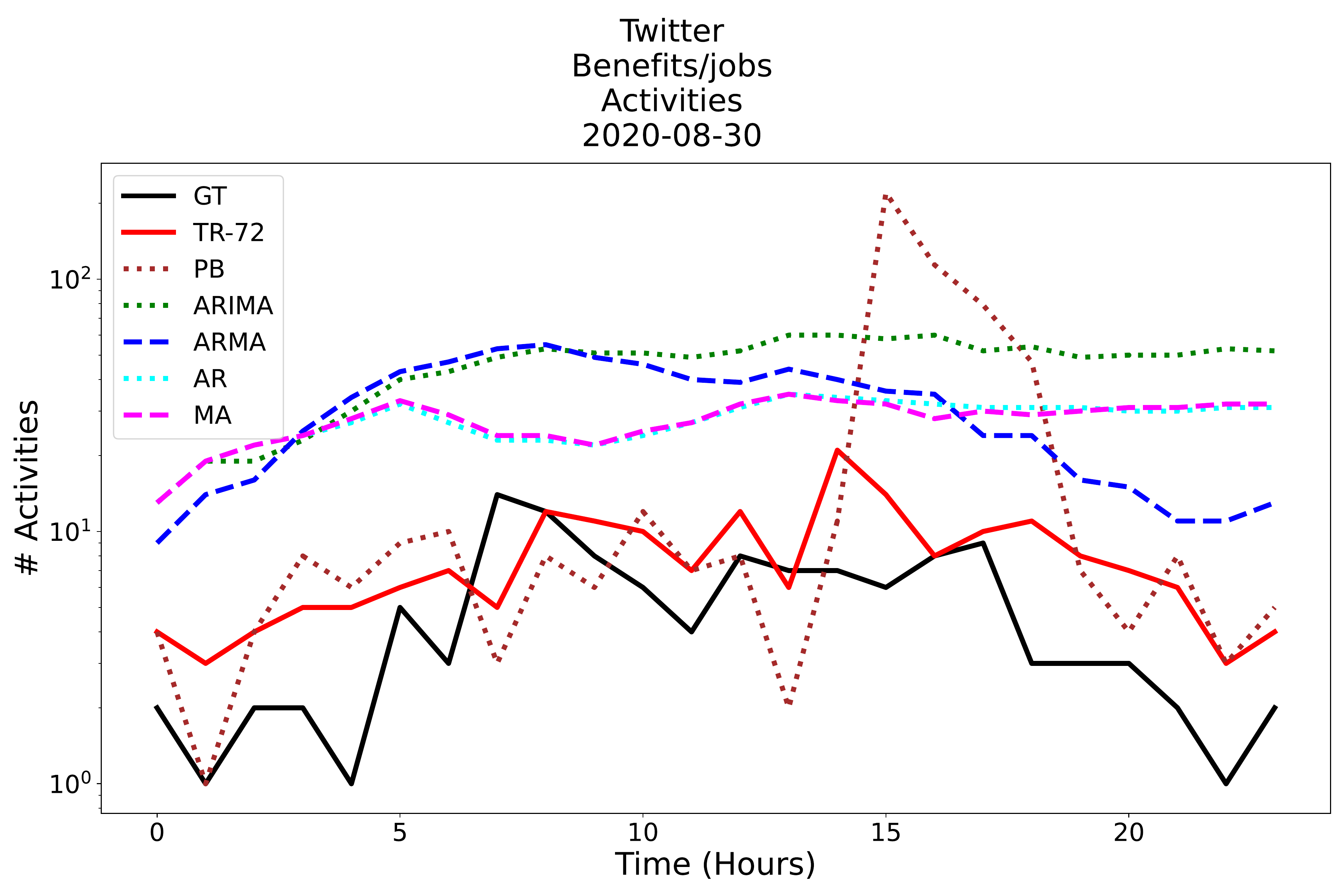} 
		    \label{fig:ts-plot2}
	}
	\\
			\subfloat[Controverseries/China/Uighur]{
         \includegraphics[scale=0.16]{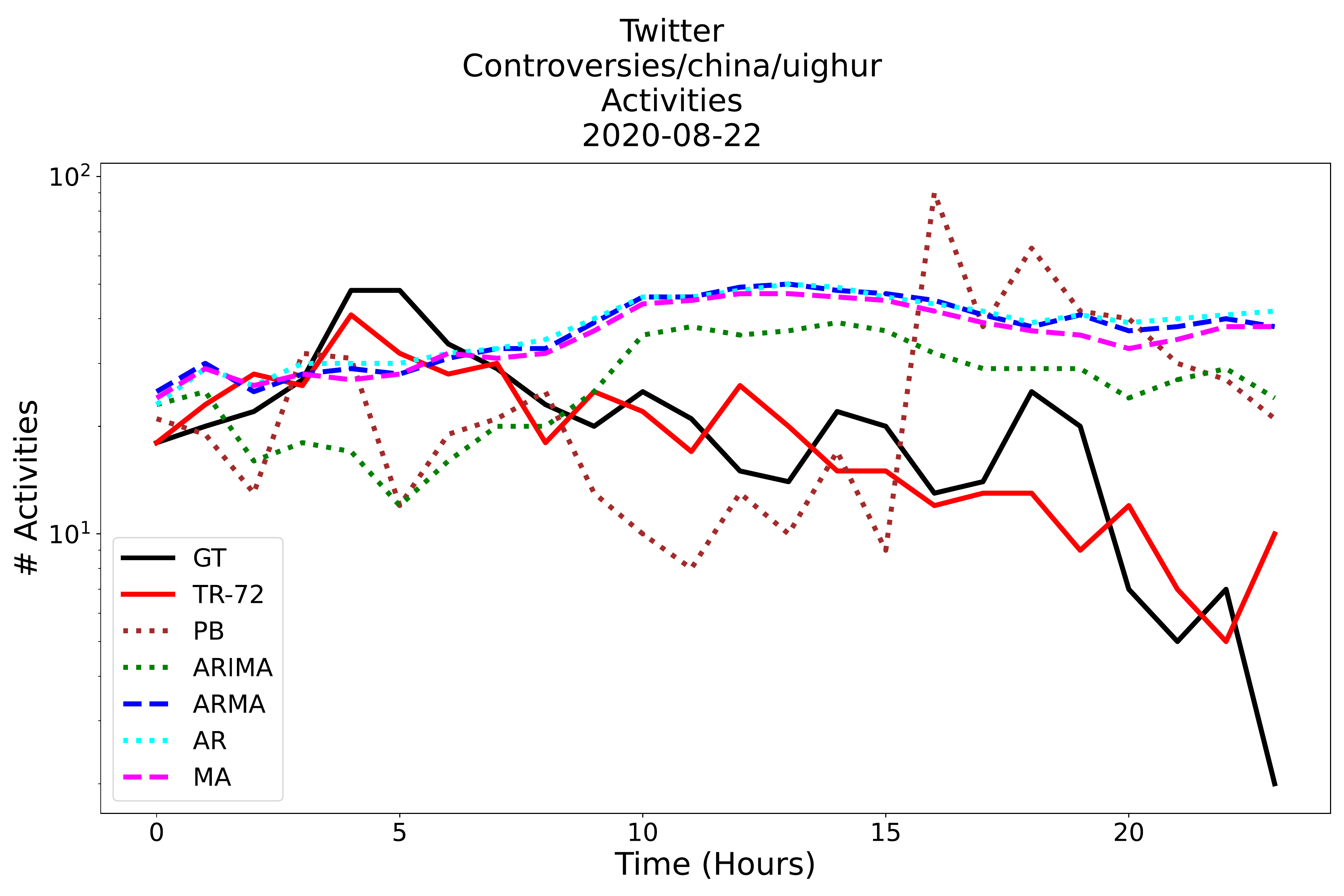} 
		    \label{fig:ts-plot3}
	}

		\subfloat[Opposition/Propoganda]{
         \includegraphics[scale=0.16]{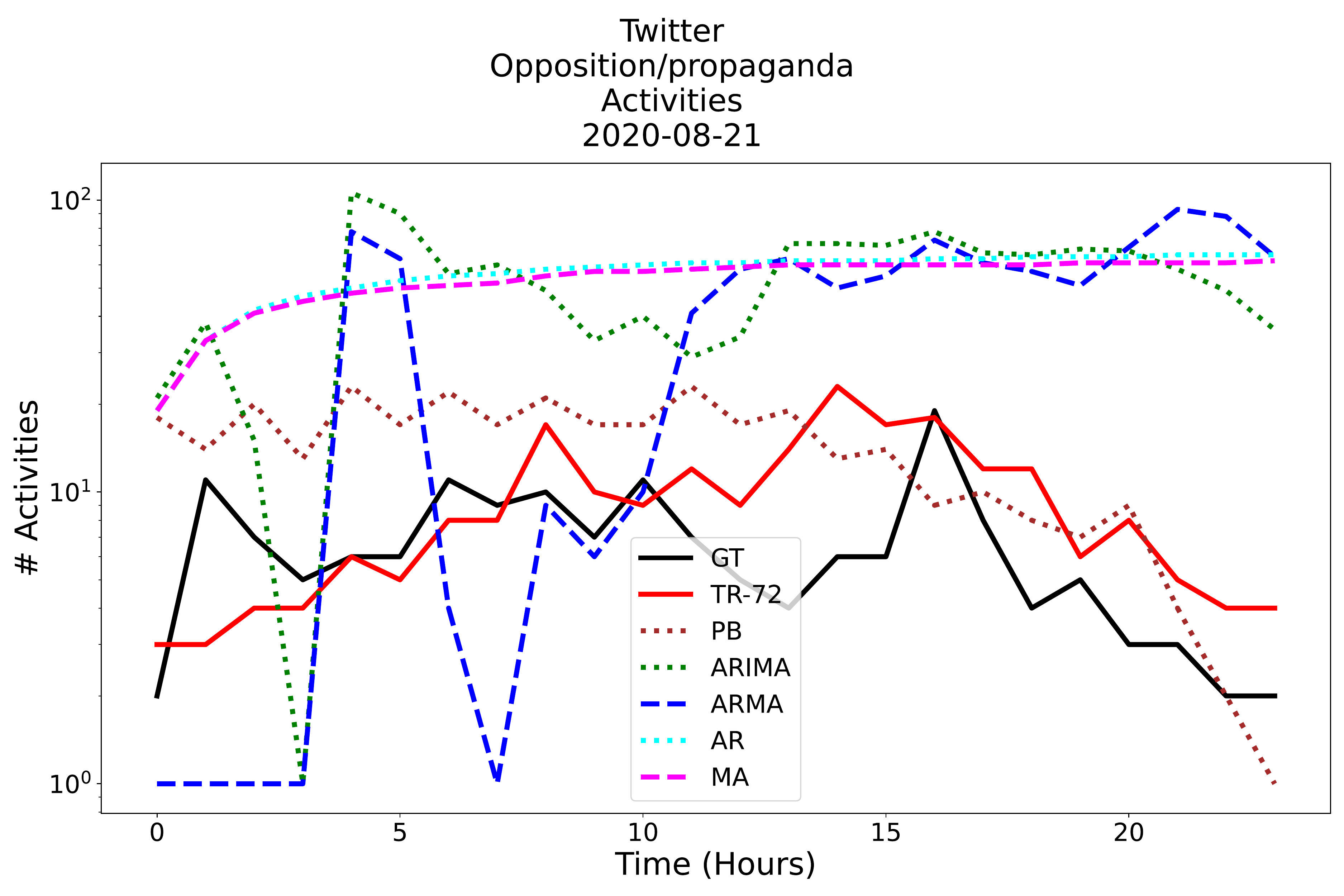} 
		    \label{fig:ts-plot4}
	}

\end{tabular}
\caption{These are some time series plots showing 24-hour periods in which the \textit{VAM-TR-72} model performed particularly well against the baseline. The red curves represent VAM's predictions, the black curves represent the ground truth, and the other curves represent the 5 baseline models.}
\label{fig:vam-ts-plots}
\end{figure*}

In summary, VAM outperformed ARMA for 49 out of 60 topic-metric pairs, or about 81.6\% of the time. It performed particularly well at the \say{volume-with-exact-timing} metrics (RMSE and MAE), the \say{approximate temporal-pattern metric} (NRMSE), and the \say{volatility} metrics (Volatility Error and Skewness Error). It performed decently on the \say{pure volume} metric (S-APE), but obviously not as well as the other metrics. Figure \ref{fig:vam-ts-plots} shows some time series plots of instances in which VAM-TR-72 performed particularly well against the baseline models.

\subsection{Temporal Feature Importances}

\begin{figure}[h!]
\centering

		\subfloat{

            \includegraphics[scale=0.4]{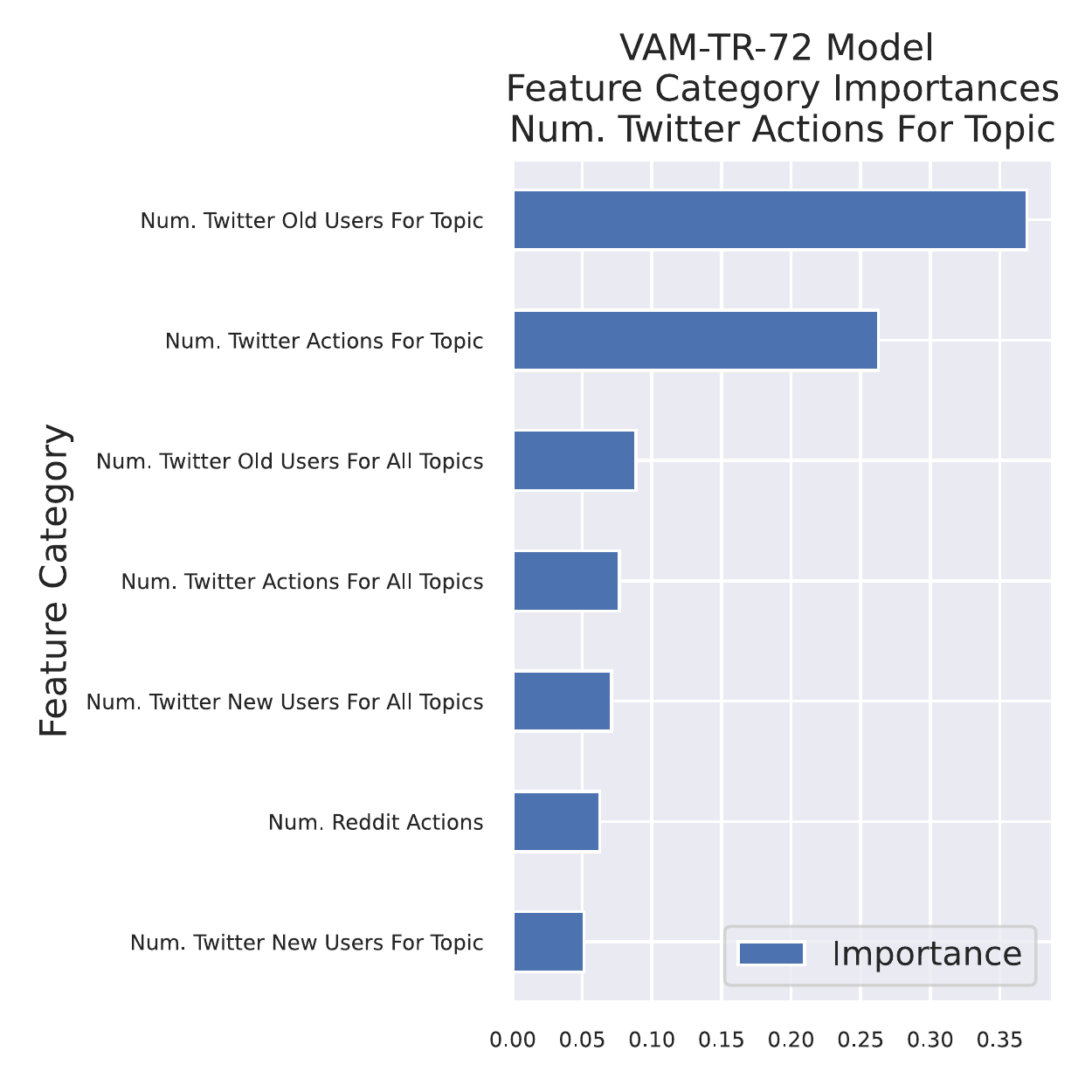} 

	}

\caption{The feature importances for the \textit{VAM-TR-72} volume prediction module.}
\label{fig:vam-twitter-ft-imps}
\end{figure}

In Figure \ref{fig:vam-twitter-ft-imps} we show a bar plot of the temporal feature importances of the XGBoost models for the \textit{number of actions} output category for the \textit{VAM-TR-72} model. The feature importances are calculated by adding up the number of times a feature is used to split the data across all trees and was calculated using the XGBoost library \cite{xgboost}. In this figure we refer to that output category as \textit{Num. Twitter Actions For Topic}.

Along the Y-axis one can see the name of each feature category. There are 6 time series feature categories, 3 for the \say{global count} time series (the ones labelled with \say{All Topics}), and 3 categories for the \say{Twitter-topic} pair time series (the ones labelled with \say{For Topic}). We normalized all the feature category importance values between 0 and 1, which is what is shown in each bar plot.

As one can see, for the \textit{VAM-T-72} model, the \textit{Num. Twitter Old Users For Topic} input time series is the most helpful for predicting the output time series \textit{Num. Twitter Actions For Topic}. In other words, according to this plot, if one wished to predict the number of actions for the topic \textit{benefits/jobs} (for example) at some future timestep, the most useful input time series would be the number of old user time series for \textit{benefits/jobs}. 

In second place in terms of importance, is the feature category \textit{Num. Twitter Activity Users For Topic}, and in third place is the feature category \textit{Num. Twitter Old Users For All Topics}. The \textit{Num. Reddit Actions} was the 5th most important feature category, ahead of \textit{Num. Twitter New users For Topic}.

\begin{table*}[]
\centering
\begin{tabular}{|cccccccccc|}
\hline
\multicolumn{10}{|c|}{\textbf{VAM and Baseline Volume Prediction Results}} \\ \hline
\multicolumn{1}{|c|}{\textbf{Rank}} &
  \multicolumn{1}{c|}{\textbf{Model}} &
  \multicolumn{1}{c|}{\textbf{RMSE}} &
  \multicolumn{1}{c|}{\textbf{MAE}} &
  \multicolumn{1}{c|}{\textbf{VE}} &
  \multicolumn{1}{c|}{\textbf{SkE}} &
  \multicolumn{1}{c|}{\textbf{S-APE}} &
  \multicolumn{1}{c|}{\textbf{NC-RMSE}} &
  \multicolumn{1}{c|}{\textbf{\begin{tabular}[c]{@{}c@{}}Overall \\ Normalized \\ Metric \\ Error\end{tabular}}} &
  \textbf{\begin{tabular}[c]{@{}c@{}}ONME \\ PIFBB \\ (\%)\end{tabular}} \\ \hline
\multicolumn{1}{|c|}{\textbf{1}} &
  \multicolumn{1}{c|}{\textbf{VAM-TR-72}} &
  \multicolumn{1}{c|}{63.7693} &
  \multicolumn{1}{c|}{45.77} &
  \multicolumn{1}{c|}{35.8454} &
  \multicolumn{1}{c|}{1.0726} &
  \multicolumn{1}{c|}{37.9726} &
  \multicolumn{1}{c|}{0.1253} &
  \multicolumn{1}{c|}{\textbf{0.0539}} &
  \textbf{16.9263} \\ \hline
\multicolumn{1}{|c|}{2} &
  \multicolumn{1}{c|}{VAM-TY-72} &
  \multicolumn{1}{c|}{65.7877} &
  \multicolumn{1}{c|}{47.0955} &
  \multicolumn{1}{c|}{35.0636} &
  \multicolumn{1}{c|}{0.9546} &
  \multicolumn{1}{c|}{37.6423} &
  \multicolumn{1}{c|}{0.1353} &
  \multicolumn{1}{c|}{0.054} &
  16.7831 \\ \hline
\multicolumn{1}{|c|}{3} &
  \multicolumn{1}{c|}{VAM-TRY-48} &
  \multicolumn{1}{c|}{66.2068} &
  \multicolumn{1}{c|}{47.181} &
  \multicolumn{1}{c|}{34.2751} &
  \multicolumn{1}{c|}{1.0059} &
  \multicolumn{1}{c|}{37.5466} &
  \multicolumn{1}{c|}{0.1322} &
  \multicolumn{1}{c|}{0.054} &
  16.7577 \\ \hline
\multicolumn{1}{|c|}{4} &
  \multicolumn{1}{c|}{VAM-TR-48} &
  \multicolumn{1}{c|}{66.5991} &
  \multicolumn{1}{c|}{47.284} &
  \multicolumn{1}{c|}{36.5073} &
  \multicolumn{1}{c|}{1.0468} &
  \multicolumn{1}{c|}{37.6235} &
  \multicolumn{1}{c|}{0.1283} &
  \multicolumn{1}{c|}{0.0547} &
  15.676 \\ \hline
\multicolumn{1}{|c|}{5} &
  \multicolumn{1}{c|}{VAM-TRY-72} &
  \multicolumn{1}{c|}{64.0651} &
  \multicolumn{1}{c|}{45.8827} &
  \multicolumn{1}{c|}{35.7631} &
  \multicolumn{1}{c|}{1.1476} &
  \multicolumn{1}{c|}{38.2717} &
  \multicolumn{1}{c|}{0.1287} &
  \multicolumn{1}{c|}{0.0548} &
  15.4899 \\ \hline
\multicolumn{1}{|c|}{6} &
  \multicolumn{1}{c|}{VAM-T-72} &
  \multicolumn{1}{c|}{63.5627} &
  \multicolumn{1}{c|}{45.6483} &
  \multicolumn{1}{c|}{36.9797} &
  \multicolumn{1}{c|}{1.1979} &
  \multicolumn{1}{c|}{37.9557} &
  \multicolumn{1}{c|}{0.127} &
  \multicolumn{1}{c|}{0.0552} &
  14.9155 \\ \hline
\multicolumn{1}{|c|}{7} &
  \multicolumn{1}{c|}{VAM-TY-48} &
  \multicolumn{1}{c|}{66.5644} &
  \multicolumn{1}{c|}{47.7133} &
  \multicolumn{1}{c|}{35.566} &
  \multicolumn{1}{c|}{1.0358} &
  \multicolumn{1}{c|}{38.5408} &
  \multicolumn{1}{c|}{0.137} &
  \multicolumn{1}{c|}{0.0553} &
  14.7477 \\ \hline
\multicolumn{1}{|c|}{8} &
  \multicolumn{1}{c|}{VAM-T-48} &
  \multicolumn{1}{c|}{64.0599} &
  \multicolumn{1}{c|}{46.0157} &
  \multicolumn{1}{c|}{36.7059} &
  \multicolumn{1}{c|}{1.1699} &
  \multicolumn{1}{c|}{38.338} &
  \multicolumn{1}{c|}{0.1292} &
  \multicolumn{1}{c|}{0.0553} &
  14.7338 \\ \hline
\multicolumn{1}{|c|}{9} &
  \multicolumn{1}{c|}{VAM-TR-24} &
  \multicolumn{1}{c|}{64.8761} &
  \multicolumn{1}{c|}{47.0618} &
  \multicolumn{1}{c|}{38.0413} &
  \multicolumn{1}{c|}{1.1434} &
  \multicolumn{1}{c|}{40.2569} &
  \multicolumn{1}{c|}{0.1235} &
  \multicolumn{1}{c|}{0.0558} &
  13.9192 \\ \hline
\multicolumn{1}{|c|}{10} &
  \multicolumn{1}{c|}{VAM-TRY-24} &
  \multicolumn{1}{c|}{65.1584} &
  \multicolumn{1}{c|}{47.2283} &
  \multicolumn{1}{c|}{37.6308} &
  \multicolumn{1}{c|}{1.1141} &
  \multicolumn{1}{c|}{41.0268} &
  \multicolumn{1}{c|}{0.1289} &
  \multicolumn{1}{c|}{0.0561} &
  13.4203 \\ \hline
\multicolumn{1}{|c|}{11} &
  \multicolumn{1}{c|}{VAM-TY-24} &
  \multicolumn{1}{c|}{65.7277} &
  \multicolumn{1}{c|}{47.6674} &
  \multicolumn{1}{c|}{37.7781} &
  \multicolumn{1}{c|}{1.1192} &
  \multicolumn{1}{c|}{40.9077} &
  \multicolumn{1}{c|}{0.1311} &
  \multicolumn{1}{c|}{0.0565} &
  12.8468 \\ \hline
\multicolumn{1}{|c|}{12} &
  \multicolumn{1}{c|}{VAM-T-24} &
  \multicolumn{1}{c|}{65.3621} &
  \multicolumn{1}{c|}{47.4219} &
  \multicolumn{1}{c|}{38.0849} &
  \multicolumn{1}{c|}{1.1958} &
  \multicolumn{1}{c|}{40.079} &
  \multicolumn{1}{c|}{0.1289} &
  \multicolumn{1}{c|}{0.0567} &
  12.5483 \\ \hline
\multicolumn{1}{|c|}{13} &
  \multicolumn{1}{c|}{ARMA} &
  \multicolumn{1}{c|}{72.5972} &
  \multicolumn{1}{c|}{55.1047} &
  \multicolumn{1}{c|}{39.4702} &
  \multicolumn{1}{c|}{1.6593} &
  \multicolumn{1}{c|}{42.9807} &
  \multicolumn{1}{c|}{0.143} &
  \multicolumn{1}{c|}{0.0648} &
  0.0 \\ \hline
\multicolumn{1}{|c|}{14} &
  \multicolumn{1}{c|}{PB} &
  \multicolumn{1}{c|}{85.6484} &
  \multicolumn{1}{c|}{61.8182} &
  \multicolumn{1}{c|}{42.5324} &
  \multicolumn{1}{c|}{0.9628} &
  \multicolumn{1}{c|}{38.525} &
  \multicolumn{1}{c|}{0.1764} &
  \multicolumn{1}{c|}{0.0649} &
  -0.0431 \\ \hline
\multicolumn{1}{|c|}{15} &
  \multicolumn{1}{c|}{ARIMA} &
  \multicolumn{1}{c|}{71.414} &
  \multicolumn{1}{c|}{54.6664} &
  \multicolumn{1}{c|}{38.9028} &
  \multicolumn{1}{c|}{1.8779} &
  \multicolumn{1}{c|}{45.0738} &
  \multicolumn{1}{c|}{0.1602} &
  \multicolumn{1}{c|}{0.0678} &
  -4.6265 \\ \hline
\multicolumn{1}{|c|}{16} &
  \multicolumn{1}{c|}{AR} &
  \multicolumn{1}{c|}{71.0034} &
  \multicolumn{1}{c|}{54.5305} &
  \multicolumn{1}{c|}{39.9082} &
  \multicolumn{1}{c|}{2.2177} &
  \multicolumn{1}{c|}{42.3746} &
  \multicolumn{1}{c|}{0.1393} &
  \multicolumn{1}{c|}{0.0684} &
  -5.5195 \\ \hline
\multicolumn{1}{|c|}{17} &
  \multicolumn{1}{c|}{MA} &
  \multicolumn{1}{c|}{79.2893} &
  \multicolumn{1}{c|}{61.0594} &
  \multicolumn{1}{c|}{42.7798} &
  \multicolumn{1}{c|}{2.1334} &
  \multicolumn{1}{c|}{44.4995} &
  \multicolumn{1}{c|}{0.1432} &
  \multicolumn{1}{c|}{0.0718} &
  -10.7239 \\ \hline
\end{tabular}
\caption{VAM and Baseline Volume Prediction Results}
\label{tab:vam-vol-ranks}
\end{table*}

\section{USER-ASSIGNMENT METHODOLOGY}

In the following subsections, we shift our focus to the User Assignment module of VAM.

\subsection{User-Assignment Lookback Factor}

Similar to how the \textit{Volume Prediction } modules utilized lookback factors (\( L^{vol}\)), we also utilized a lookback factor parameter for the \textit{User-Assignment} task, \(L^{user}\). We set this value to 24 hours.  So, in other words, VAM's user-assignment module only uses the past 24 hours of user interaction history when making predictions. The assumption here is that recent user-interaction history is all that is needed to make accurate user-to-user predictions. We call this new truncated version of the temporal sequence of graphs, \(G\), \(G^{recent}\). Using this information we now describe the user-assignment algorithm \cite{Mubang-VAM-IEEE-TRANS}.

\subsection{User Assignment Explained}

A recent history table called \(H^{recent}\) is created from the history sequence of graphs, \(G^{recent}\). This table contains event records, with each record being defined as a tuple containing (1) the timestamp, (2) the name of the child user, (3) the name of the parent user, (4) the number of interactions between the two users, (5) a flag indicating if the child user is new, and (6) a flag indicating if the parent user is new.

Using this table and the volume count of old users from the \textit{Volume Prediction}, module, \textit{VAM} utilizes weighted random sampling to predict the set of active old users at T + 1, \(\hat{O}^{T + 1} \). Using the new user volume prediction counts, \textit{VAM} is also able to create the set of active new users at T + 1,  \(\hat{N}^{T + 1} \). Multiple data structures for each set of users are used to keep track of 4 main user attributes: (1) the user's probability of activity, (2) the user's probability of influence, (3) the user's list of parents it is most likely to interact with,  and (4) the probability a user would interact with each parent in their respective parent list. 

It is easy to obtain these 4 attributes for the old users because their history is available in the \(H^{recent}\) table. However, for new users, \textit{VAM} must infer what their attributes would most likely be. In order to do this, \textit{VAM} uses a \textit{User Archetype Table}, which is created with the use of a random sampling algorithm applied to the set of old users in the \(H^{recent}\) table. The assumption  is that new users in the future are likely to have the same attributes as old users in the recent past. 

\textit{VAM} then uses weighted random sampling in order to assign edges among the users in the \(\hat{O}^{T + 1} \) and \(\hat{N}^{T + 1} \) set. \textit{VAM} \say{knows} how many total actions to assign among all users because the activity volume time series was predicted in the \textit{Volume-Prediction} task. The final set of nodes and edges predicted at \(T+1\) is known as \(G^{future}_1\). \textit{VAM} updates the history table \(H^{recent}\) with the new graph \(G^{future}_1\), and then repeats the process of predicting old users, new users, and user-user interactions until it has predicted the full sequence \(G^{future} = \{G^{future}_1, G^{future}_2, ... G^{future}_S\}\). The supplemental materials \cite{vam-suppl} contain a visual representation of the User Assignment algorithm. For more details, see \cite{Mubang-VAM-IEEE-TRANS}.

\section{User Assignment Results}

In this section we discuss the User Assignment results. Since the user-assignment algorithm is probabilistic, we performed 5 trials, and averaged their metric results. The supplemental materials \cite{vam-suppl} contain tables showing the standard deviations and variation coefficients of the metric results across the 5 trials.

\subsection{Measuring Old Users}

In order to measure the accuracy of the old user prediction task, the Weighted Jaccard Similarity metric was used, which is also known as the Ruzicka Similarity \cite{jacc-sim}. It was used to measure how well VAM predicted the old users in each hour, as well as how \say{influential} they were. In this case, influence is defined quantitatively as the number of retweets, replies, and quotes a user's tweets received. For more details regarding this metric, see the supplemental materials \cite{vam-suppl}.

\subsection{Measuring Old and New Users Together}

Since our task involves predicting the creation and activity of new users, in addition to  activity of old users, defining and measuring new user predictive success has complexities. The names of a new user are unknown before they appear in the ground truth. Hence, it is impossible to exactly match a new user that \textit{VAM} generates with a new user in the ground truth. So, in order to work around this issue, we measure new user prediction success using more macroscopic views of the network in the same fashion as \cite{Mubang-VAM-IEEE-TRANS}. We call these types of results, \textit{Network Structure} results. Specifically, we used the Page Rank Distribution \cite{page-rank} of the weighted indegree of the network and the Complementary Cumulative Degree Histogram (CCDH) \cite{simpson-rh-paper} of the unweighted indegree of the network. In order to measure the distance between the predicted and actual Page Rank distributions we used the Earth Mover's Distance Metric \cite{emd-rub}. In order to measure the distance between the CCHD's of the predicted network and the ground truth network,  the Relative Hausdorff (RH) Distance \cite{simpson-rh-paper} was used. 

\subsection{Baseline Used}

For the user-assignment task, we used the \textit{Persistence Baseline} as a baseline. Similar to the Volume Prediction task, it is created by shifting the user-to-user networks spanning \(T\) to \(T - S\) up to period \(T + 1\) to \(T + S\). This same baseline was also used in \cite{renhao-github-atp, github-renhao-cve2vec, anthony-wh, socialcube-ieee, 2farm-challenge-long}. 

Other works in the literature are unsuitable as baselines due to various reasons. The approaches in \cite{socialcube-ieee, renhao-github-atp, Horawalavithana2019MentionsSecurity, saadat-github-archetypes, bidoki-coding-dyn} are all prediction approaches that are platform-specific to Github, and do not translate easily to Twitter. The approaches in \cite{blythe-FARM, usc2020-massive-sims} predict Twitter user activity, but not in a way that is comparable to VAM. Those works predict user-to-tweet interactions as a classification task, whereas VAM predicts user-to-user interactions as a regression task. The general link prediction methods can predict Twitter user activity, however, they can only predict the activity of old users and not new users \cite{dg2vec, tnode, temp-lp-dunlavy, non-neg-mf, gao-temp-lp, non-para-lp-sarkar, ahmed-ts-rand-walk}. The Persistence Baseline can easily predict new users since it uses historical information for its predictions. For example, if it is known that there were 10,000 new users in the past period, the Persistence Baseline trivially predicts that there will be 10,000 new users in the future period. Lastly, in addition to not predicting new users, some works do not scale to the large networks used in this work   \cite{CHEN2019221, Hernandez2020DeepLearning, prasha-naam, dg2vec,non-para-lp-sarkar, ahmed-ts-rand-walk}. 

\subsection{Old User Prediction Results}

Table \ref{tab:vam-wjs} contains the old user prediction results using the Weighted Jaccard Similarity metric. Since this is a similarity metric, higher scores are better. As one can see in the table we refer to this model as the \textit{VAM-TR-72V-24U} model. This is a \textit{VAM} model that has a volume lookback factor (\(L^{vol})\) of 72 hours and a user-assignment lookback factor (\(L^{user})\) of 24 hours.  The numbers in bold represent the best results. As one can see in this table, VAM outperformed the Persistence Baseline on 8 out of 10 topics. VAM performed particularly well on the \textit{benefits/development/roads}, \textit{leadership/sharif}, and \textit{controversies/china/uighur} topics. The Percent Improvement From Baseline (PIFB) scores on these topics are about 220\%, 214\%, and 120\%, respectively. 

\subsection{Network Structure Results}

Table \ref{tab:cp5-emd-vam} shows the results for the Earth Mover's Distance metric (lower is better).  As one can see, \textit{VAM} outperformed the baseline on this metric for 8 out of 10 topics. Note, that the two topics where performance is less than the baseline have the least activity. \textit{VAM} performed particularly well for the \textit{controversies/pakistan/baloch}, \textit{benefits/development/roads}, and \textit{opposition/propoganda} topics. The Percent Improvement From Baseline (PIFB) scores for those topics were 27.29\%, 20.89\%, and 19.2\%, respectively.  

Table \ref{tab:cp5-rhd-vam} shows \textit{VAM}'s Relative Hausdorff Distance results (lower is better). Similar to the Earth Mover's Distance results, VAM beat the baseline on 8 out of 10 topics. It performed particularly well on the \textit{controversies/pakistan/baloch}, \textit{leadership/sharif}, and \textit{controversies/pakistan/students} topics. The percent improvement scores for those topics were 25.24\%, 22.8\%, and 18.94\%, respectively.

\begin{table}[]
\centering
\begin{tabular}{|cccc|}
\hline
\multicolumn{4}{|c|}{\textbf{\begin{tabular}[c]{@{}c@{}}VAM-TR-72V-24U Weighted\\ Jaccard Similarity Results\end{tabular}}}                             \\ \hline
\multicolumn{1}{|c|}{\textbf{Topic}} &
  \multicolumn{1}{c|}{\textbf{\begin{tabular}[c]{@{}c@{}}VAM-TR-\\ 72V-24U\end{tabular}}} &
  \multicolumn{1}{c|}{\textbf{\begin{tabular}[c]{@{}c@{}}Persistence \\ Baseline\end{tabular}}} &
  \textbf{\begin{tabular}[c]{@{}c@{}}PIFB\\ (\%)\end{tabular}} \\ \hline
\multicolumn{1}{|c|}{benefits/development/roads}      & \multicolumn{1}{c|}{\textbf{0.1192}} & \multicolumn{1}{c|}{0.0373}          & \textbf{219.7479} \\ \hline
\multicolumn{1}{|c|}{leadership/sharif}               & \multicolumn{1}{c|}{\textbf{0.1352}} & \multicolumn{1}{c|}{0.043}           & \textbf{214.1777} \\ \hline
\multicolumn{1}{|c|}{controversies/china/uighur}      & \multicolumn{1}{c|}{\textbf{0.1621}} & \multicolumn{1}{c|}{0.0738}          & \textbf{119.6832} \\ \hline
\multicolumn{1}{|c|}{controversies/pakistan/baloch}   & \multicolumn{1}{c|}{\textbf{0.0567}} & \multicolumn{1}{c|}{0.0308}          & \textbf{83.8803}  \\ \hline
\multicolumn{1}{|c|}{opposition/propaganda}           & \multicolumn{1}{c|}{\textbf{0.0958}} & \multicolumn{1}{c|}{0.056}           & \textbf{71.1109}  \\ \hline
\multicolumn{1}{|c|}{benefits/development/energy}     & \multicolumn{1}{c|}{\textbf{0.0744}} & \multicolumn{1}{c|}{0.0455}          & \textbf{63.4634}  \\ \hline
\multicolumn{1}{|c|}{controversies/china/border}      & \multicolumn{1}{c|}{\textbf{0.0851}} & \multicolumn{1}{c|}{0.0572}          & \textbf{48.827}   \\ \hline
\multicolumn{1}{|c|}{leadership/bajwa}                & \multicolumn{1}{c|}{\textbf{0.1008}} & \multicolumn{1}{c|}{0.0878}          & \textbf{14.8343}  \\ \hline
\multicolumn{1}{|c|}{benefits/jobs}                   & \multicolumn{1}{c|}{0.068}           & \multicolumn{1}{c|}{\textbf{0.0688}} & -1.1927           \\ \hline
\multicolumn{1}{|c|}{controversies/pakistan/students} & \multicolumn{1}{c|}{0.062}           & \multicolumn{1}{c|}{\textbf{0.1118}} & -44.586           \\ \hline
\end{tabular}
\caption{VAM-TR-72V-24U Weighted Jaccard Similarity Results}
\label{tab:vam-wjs}
\end{table}

\begin{table}[]
\begin{tabular}{|cccc|}
\hline
\multicolumn{4}{|c|}{\textbf{\begin{tabular}[c]{@{}c@{}}VAM-TR-72V-24U \\ Earth Mover's Distance Results\end{tabular}}}                              \\ \hline
\multicolumn{1}{|c|}{\textbf{Topic}} &
  \multicolumn{1}{c|}{\textbf{\begin{tabular}[c]{@{}c@{}}VAM-TR-\\ 72V-24U\end{tabular}}} &
  \multicolumn{1}{c|}{\textbf{PB}} &
  \textbf{PIFB (\%)} \\ \hline
\multicolumn{1}{|c|}{controversies/pakistan/baloch}   & \multicolumn{1}{c|}{\textbf{0.0358}} & \multicolumn{1}{c|}{0.0492}          & \textbf{27.29} \\ \hline
\multicolumn{1}{|c|}{benefits/development/roads}      & \multicolumn{1}{c|}{\textbf{0.1076}} & \multicolumn{1}{c|}{0.136}           & \textbf{20.89} \\ \hline
\multicolumn{1}{|c|}{opposition/propaganda}           & \multicolumn{1}{c|}{\textbf{0.0963}} & \multicolumn{1}{c|}{0.1192}          & \textbf{19.2}  \\ \hline
\multicolumn{1}{|c|}{leadership/sharif}               & \multicolumn{1}{c|}{\textbf{0.082}}  & \multicolumn{1}{c|}{0.0965}          & \textbf{14.95} \\ \hline
\multicolumn{1}{|c|}{controversies/china/border}      & \multicolumn{1}{c|}{\textbf{0.1144}} & \multicolumn{1}{c|}{0.1276}          & \textbf{10.41} \\ \hline
\multicolumn{1}{|c|}{controversies/china/uighur}      & \multicolumn{1}{c|}{\textbf{0.1137}} & \multicolumn{1}{c|}{0.1233}          & \textbf{7.79}  \\ \hline
\multicolumn{1}{|c|}{benefits/development/energy}     & \multicolumn{1}{c|}{\textbf{0.1896}} & \multicolumn{1}{c|}{0.1979}          & \textbf{4.22}  \\ \hline
\multicolumn{1}{|c|}{leadership/bajwa}                & \multicolumn{1}{c|}{\textbf{0.1971}} & \multicolumn{1}{c|}{0.2038}          & \textbf{3.27}  \\ \hline
\multicolumn{1}{|c|}{benefits/jobs}                   & \multicolumn{1}{c|}{0.2137}          & \multicolumn{1}{c|}{\textbf{0.2087}} & -2.41          \\ \hline
\multicolumn{1}{|c|}{controversies/pakistan/students} & \multicolumn{1}{c|}{0.1945}          & \multicolumn{1}{c|}{\textbf{0.1669}} & -16.48         \\ \hline
\end{tabular}
\caption{VAM-TR-72V-24U Earth Mover's Distance Results}
\label{tab:cp5-emd-vam}
\end{table}

\begin{table}[]
\begin{tabular}{|cccc|}
\hline
\multicolumn{4}{|c|}{\textbf{\begin{tabular}[c]{@{}c@{}}VAM-TR-72V-24U \\ Relative Hausdorff Distance Results\end{tabular}}}                         \\ \hline
\multicolumn{1}{|c|}{\textbf{Topic}} &
  \multicolumn{1}{c|}{\textbf{\begin{tabular}[c]{@{}c@{}}VAM-TR-\\ 72V-24U\end{tabular}}} &
  \multicolumn{1}{c|}{\textbf{PB}} &
  \textbf{PIFB (\%)} \\ \hline
\multicolumn{1}{|c|}{controversies/pakistan/baloch}   & \multicolumn{1}{c|}{\textbf{0.9015}} & \multicolumn{1}{c|}{1.2059}          & \textbf{25.24} \\ \hline
\multicolumn{1}{|c|}{leadership/sharif}               & \multicolumn{1}{c|}{\textbf{0.7985}} & \multicolumn{1}{c|}{1.0344}          & \textbf{22.8}  \\ \hline
\multicolumn{1}{|c|}{controversies/pakistan/students} & \multicolumn{1}{c|}{\textbf{0.6138}} & \multicolumn{1}{c|}{0.7572}          & \textbf{18.94} \\ \hline
\multicolumn{1}{|c|}{benefits/development/roads}      & \multicolumn{1}{c|}{\textbf{0.7651}} & \multicolumn{1}{c|}{0.8891}          & \textbf{13.94} \\ \hline
\multicolumn{1}{|c|}{benefits/jobs}                   & \multicolumn{1}{c|}{\textbf{0.6669}} & \multicolumn{1}{c|}{0.7355}          & \textbf{9.32}  \\ \hline
\multicolumn{1}{|c|}{benefits/development/energy}     & \multicolumn{1}{c|}{\textbf{0.688}}  & \multicolumn{1}{c|}{0.7253}          & \textbf{5.14}  \\ \hline
\multicolumn{1}{|c|}{controversies/china/uighur}      & \multicolumn{1}{c|}{\textbf{0.7696}} & \multicolumn{1}{c|}{0.8018}          & \textbf{4.01}  \\ \hline
\multicolumn{1}{|c|}{leadership/bajwa}                & \multicolumn{1}{c|}{\textbf{1.0904}} & \multicolumn{1}{c|}{1.0906}          & \textbf{0.01}  \\ \hline
\multicolumn{1}{|c|}{controversies/china/border}      & \multicolumn{1}{c|}{0.9512}          & \multicolumn{1}{c|}{\textbf{0.9411}} & -1.08          \\ \hline
\multicolumn{1}{|c|}{opposition/propaganda}           & \multicolumn{1}{c|}{1.2339}          & \multicolumn{1}{c|}{\textbf{1.2017}} & -2.68          \\ \hline
\end{tabular}
\caption{VAM-TR-72V-24U Relative Hausdorff Distance}
\label{tab:cp5-rhd-vam}
\end{table}

\subsection{Hardware and Runtime Information}

The 5 trials were run in parallel across 5 computers, each with an Intel Xeon E5-260 v4 CPU. Each CPU was comprised of 2 sockets, 8 cores, and 16 threads. Each computer had 128 GB of memory. The average runtime across all 5 trials was approximately 27 minutes.

\section{CONCLUSION}

In this work, we discussed the \textit{VAM} simulator \cite{Mubang-VAM-IEEE-TRANS}, an end-to-end approach for time series prediction and temporal link prediction and applied it to the CPEC Twitter dataset. We showed that VAM could outperform the Persistence Baseline, ARIMA, ARMA, AR, and MA models on the \textit{Volume Prediction} task. We then showed that VAM could outperform the Persistence Baseline on the User Assignment task. 
On the Volume-Prediction task, VAM outperformed the best baseline model (ARMA) on 49 out of 60 (or 81.6\%) of all topic-metric pairs. Furthermore, we showed that external Reddit and YouTube features aid VAM with the Volume Prediction task. 

For the User-Assignment task, VAM outperformed the Persistence Baseline on 24 out of 30 (or 80\%) of all topic-metric pairs. Also, we showed that VAM can predict the creation of new users, unlike many previous link prediction approaches that only focus on the prediction of old user-to-user interactions. Furthermore, we explained that VAM's user-assignment is quite fast, taking only 27 minutes to simulate the activity of millions of user-to-user edges. 

By showing VAM's strong performance on the CPEC dataset, we lend more credence to the notion that VAM can serve as a general social media simulator, and not one that is just specific to the Venezuelan Political dataset \cite{Mubang-VAM-IEEE-TRANS}. Future work  involves utilizing a machine-learning model for the \textit{User-Assignment} module, as well as trying LSTM neural networks for both the \textit{Volume Prediction} and \textit{User-Assignment} modules.

\section*{ACKNOWLEDGMENT}

The authors thank Leidos for providing the Twitter, YouTube, and Reddit data. 
This work is partially supported by DARPA and Air Force Research Laboratory via contract FA8650-18-C-7825.


\bibliographystyle{IEEEtran}
\bibliography{ref}

\end{document}